\documentclass[11pt, letterpaper, logo, onecolumn, numbering]{googledeepmind}

\usepackage{times}

\usepackage[numbers, sort&compress]{natbib}
\usepackage{multicol}
\usepackage{hyperref}
\usepackage{amsmath}
\usepackage{amsfonts}
\usepackage{graphicx}  
\usepackage{subcaption}
\usepackage[font=small,labelfont=bf]{caption}
\usepackage{array}

\usepackage{xcolor}
\usepackage{xspace}
\usepackage{algorithm}
\usepackage{algpseudocode}
\usepackage{booktabs}
\usepackage{balance}

\usepackage{mdframed}
\usepackage[many]{tcolorbox}
\usepackage{lipsum}

\definecolor{myboxcolor}{RGB}{255,242,230}

\newcommand{\redit}{\texttt{Robo\textsc{ART}}\xspace}

\usepackage{mathtools}

\newcommand{\nonhybridpolicy}[0]{diffusion policy\xspace}

\def\shownotes{1}  
\ifnum\shownotes=1
\newcommand{\authnote}[2]{{$\ll$\textsf{\footnotesize #1 notes: #2}$\gg$}}
\else
\newcommand{\authnote}[2]{}
\fi

\title{Predictive Red Teaming: \\ Breaking Policies Without Breaking Robots}

\correspondingauthor{Anirudha Majumdar}

\author[  \hspace{-1ex}]{Anirudha Majumdar$^{\dagger, 1,2}$, Mohit Sharma$^{1}$, Dmitry Kalashnikov$^{1}$, \\ Sumeet Singh$^{1}$, Pierre Sermanet$^{1}$, Vikas Sindhwani$^{1}$}
\affil[ \hspace{-1ex}]{$^{1}$Google DeepMind, $^{2}$ Princeton University}

\begin{abstract}
\vspace{-18pt}
{\normalfont Webpage}: \href{https://predictive-red-team.github.io/}{predictive-red-team.github.io}
\vspace{8pt}

Visuomotor policies trained via imitation learning are capable of performing challenging manipulation tasks, but are often extremely brittle to lighting, visual distractors, and object locations.
These vulnerabilities can depend unpredictably on the specifics of training, and are challenging to expose without time-consuming and expensive hardware evaluations. We propose the problem of \emph{predictive red teaming}: discovering vulnerabilities of a policy with respect to environmental factors, and predicting the corresponding performance degradation \emph{without} hardware evaluations in off-nominal scenarios. In order to achieve this, we develop \redit: an automated red teaming (ART) pipeline that (1) modifies nominal observations using generative image editing to vary different environmental factors, and (2) predicts performance under each variation using a policy-specific anomaly detector executed on edited observations. Experiments across 500+ hardware trials in twelve off-nominal conditions for visuomotor diffusion policies demonstrate that \redit~predicts performance degradation with high accuracy (less than $0.19$ average difference between predicted and real success rates). 
We also demonstrate how predictive red teaming enables \emph{targeted data collection}: fine-tuning with data collected under conditions predicted to be adverse boosts baseline performance by 2--7x.
\end{abstract}

\begin{document}
\maketitle

\begin{center}
\vspace{5pt}
    \centering
    \captionsetup{type=figure}
    \includegraphics[width=\textwidth]{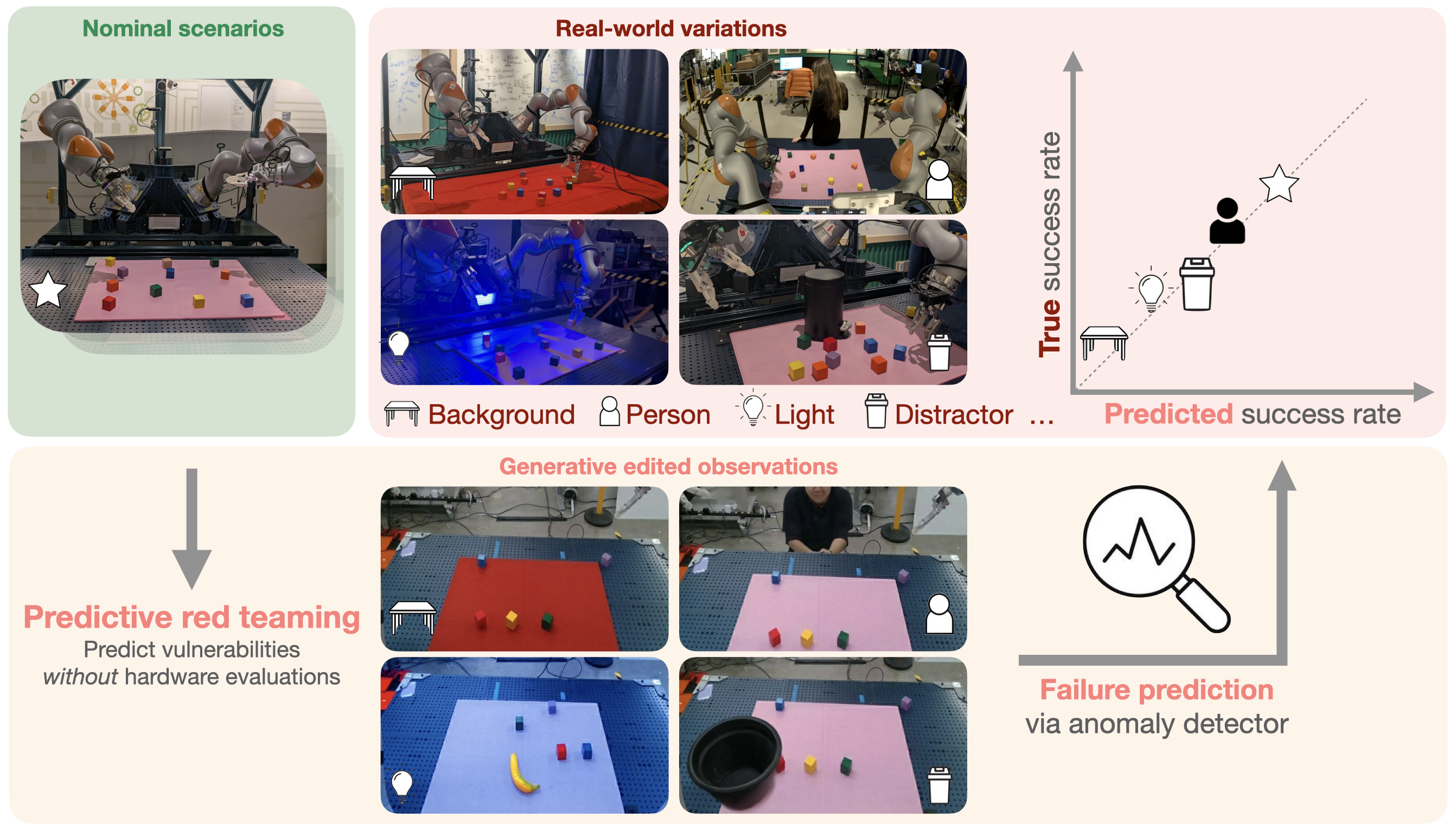}
    \captionof{figure}{We propose \emph{predictive red teaming}: discovering vulnerabilities of a policy with respect to environmental factors and predicting the corresponding performance degradation \emph{without} hardware evaluations in off-nominal scenarios. Our approach modifies nominal observations using generative image editing to reflect changes in environmental factors (e.g., background, lighting, injecting humans and other distractors), and predicts the resulting performance degradation via anomaly detection.  \label{fig:anchor}}
\end{center}%

\section{Introduction}
\label{sec:intro}

Is it possible to expose the vulnerabilities of a given robot policy with respect to changes in environmental factors such as lighting, visual distractors, and object placement \emph{without performing hardware evaluations in these scenarios}? As we seek to deploy robots in environments with ever-increasing complexity, it becomes imperative to develop scalable methods for predicting how well they will generalize when faced with unseen scenarios. Performing hardware evaluations to discover vulnerabilities --- which can depend in surprising ways on the specifics of policy training and architecture --- is often prohibitively expensive to set up and execute, especially when the goal is to test the limits of safe deployment in a sufficiently diverse set of scenarios. 

\begin{figure*}[t]
    \centering
    \includegraphics[width=\textwidth]{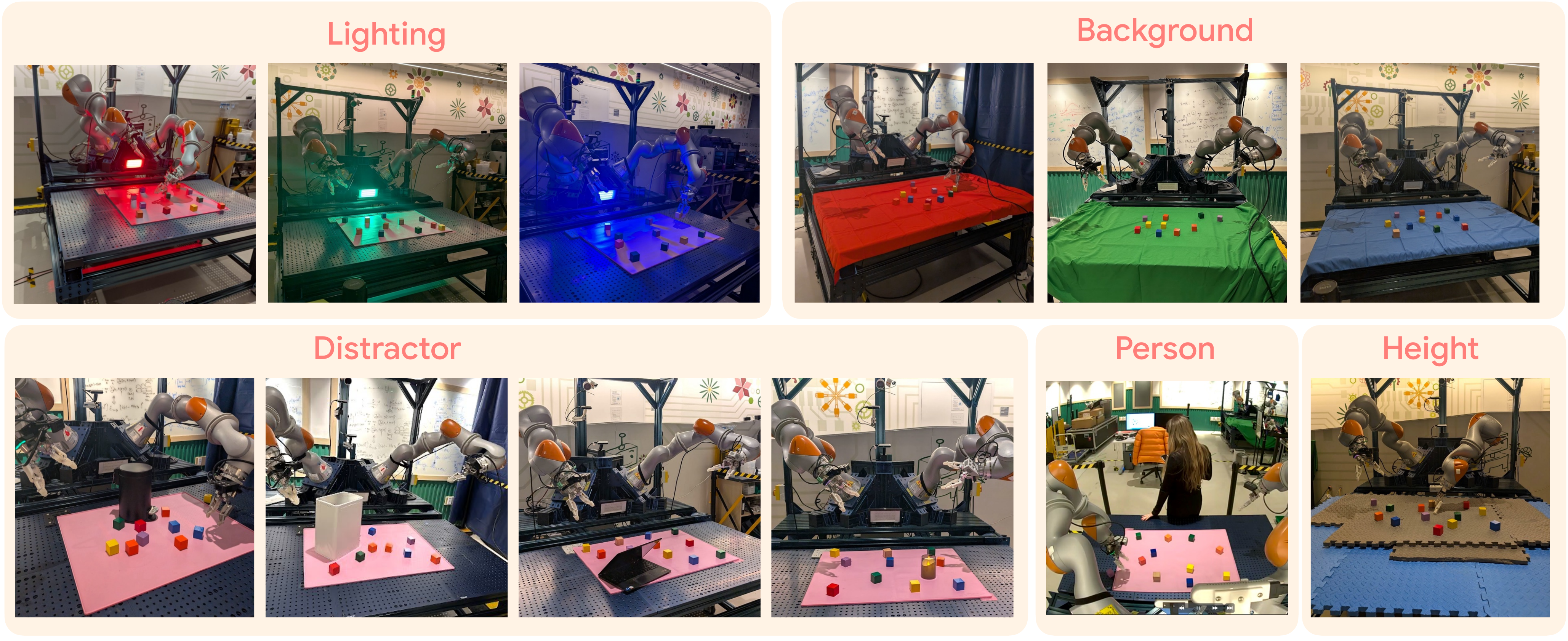}
    \caption{We evaluate \redit's predictions using 500+ hardware trials in twelve off-nominal conditions.}
    \label{fig:all factors}
    \vspace{-5pt}
\end{figure*}

As an example, consider a visuomotor diffusion policy~\cite{chi2023diffusion} trained to perform pick-and-place tasks via behavior cloning (Fig.~\ref{fig:anchor}). The policy is trained with a large dataset: over 3K+ demonstrations with varied objects, locations, and visual distractors. Will the policy generalize well to a change in the height of the table by a few centimeters (as one may plausibly predict due to the variations in 2D object locations in the training dataset) compared to when a human is standing closer to the table than seen during training? If so, what is the absolute degradation of the success rate in each case? As it turns out, the above prediction is incorrect: the success rate of the policy degrades from $\sim65\%$ under nominal conditions to $\sim10\%$ by changing the table height, and remains roughly constant with a human close to the table. Predicting the relative and absolute impact of other factors (e.g., lighting, table backgrounds, object distractors; Fig.~\ref{fig:all factors}) can be even more challenging.


\vspace{2pt}
{\bf Contribution 1 (\emph{Predictive Red Teaming}).} We introduce and formalize the problem of \emph{predictive red teaming}: discovering vulnerabilities of a given policy with respect to changes in environmental factors, and predicting the (relative or absolute) degradation in performance \emph{without} performing hardware evaluations in off-nominal scenarios. 
\vspace{2pt}

The ability to perform predictive red teaming has a number of important consequences. First, it enables \emph{targeted deployment}: by understanding the envelope of conditions that will yield satisfactory performance, we can choose where the policy is deployed. Second, it enables \emph{policy comparison}: knowing the relative vulnerabilities of different policies allows us to select one that is more likely to meet deployment needs. Third, it enables \emph{targeted data collection}: if we know that certain environmental conditions degrade performance more than others, we can re-train the policy with additional data from the adverse conditions in order to help patch vulnerabilities. 

\vspace{2pt}
{\bf Contribution 2 (\redit).} We introduce \redit --- robotics automated red teaming (ART) --- an approach to predictive red teaming for visuomotor policies based on generative image editing and anomaly detection. 
\vspace{2pt}

The pipeline for \redit~has two main steps: \emph{edit} and \emph{predict} (Fig.~\ref{fig:anchor}). First, we use automated image editing tools~\cite{baldridge2024imagen, esser2024scaling, controlnet, saharia2022photorealistic} to modify a set of nominal RGB observations by varying different factors of interest (e.g., lighting, distractors, object locations) in a fine-grained and realistic manner via language instructions (e.g., ``add a person close to the table"; Fig.~\ref{fig:anchor}). The second step is to predict the degradation in performance induced by each environmental factor using \emph{anomaly detection}. Specifically, we find that a simple anomaly detector that computes distances in policy embedding space between edited observations and a set of nominal observations (with an anomaly threshold computed using \emph{conformal prediction} \cite{vovk2005algorithmic}) is surprisingly predictive of both relative and absolute performance degradation. 


\vspace{2pt}
{\bf Contribution 3 (\emph{Demonstration for visuomotor diffusion policies}).} We evaluate \redit using 500+ hardware experiments that vary twelve environmental factors for two visuomotor diffusion policies with significantly different architectures.
\vspace{2pt}

We find that \redit~predicts performance degradation with a high degree of accuracy, e.g., correctly predicting that the changed table height will degrade performance significantly more than a human distractor. The difference between predicted and real success rates averaged across the twelve factors is 0.1 and 0.19 respectively for the two policies. 

\vspace{2pt}
{\bf Contribution 4 (\emph{Targeted data collection}).} We demonstrate the utility of predictive red teaming for targeted data collection by co-finetuning a policy with data collected in scenarios predicted to yield low performance. 
\vspace{2pt}

Co-finetuning the policy with data from the three conditions predicted to be the most adverse boosts performance in these conditions by 2--7x. Moreover, targeted data collection also yields \emph{cross-domain generalization}: the performance of the policy is improved by 2--5x even for conditions where we did not collect data.



\section{Related Work}
\label{sec:related work} 


{\bf Red teaming.} The concept of red teaming originated in the military realm, where a team posing as the enemy tries to find vulnerabilities of a military plan~\cite{zenko2015red}. In recent years, the practice of red teaming has been adopted for finding vulnerabilities of large language models (LLMs) in terms of bias, misuse, and other harmful behavior~\cite{achiam2023gpt, team2023gemini, ganguli2022red, bai2022training, wei2024jailbroken}. While red teaming for LLMs was initially performed by human evaluators, this limits the coverage of possible issues that can be discovered. As a result, recent work has sought to partially automate the process of red teaming~\cite{perez2022red, tong2024mass, zou2023universal, chao2023jailbreaking, liu2023autodan, mehrotra2023tree}, e.g., by using LLMs themselves to discover vulnerabilities. 

While there is a growing literature on red teaming for vision-language models~\cite{li2024red, liu2024arondight} and text-to-image generative models~\cite{rando2022red, gandikota2023erasing}, red teaming for robotics is still nascent. Recent work has considered \emph{embodied red teaming} for finding flaws in language-conditioned robotic foundation models~\cite{karnik2024embodied}. Specifically, \cite{karnik2024embodied} focuses on \emph{instruction generalization}: how well does a policy perform when faced with novel language instructions? As such, all evaluations in~\cite{karnik2024embodied} are performed in simulation. Related work has also considered \emph{jailbreaking} LLM-powered robots~\cite{robey2024jailbreaking}, i.e., finding adversarial prompts that override safety guardrails and cause robots to perform harmful actions. In contrast to \cite{karnik2024embodied, robey2024jailbreaking}, our focus is on finding \emph{environmental factors} (e.g., background colors, lighting, object locations) that degrade the performance of a given policy without performing hardware evaluations in off-nominal scenarios. The work in~\cite{pumacay2024colosseum} uses simulation to assess the generalization of policies with respect to environmental factors. However, setting up an accurate simulator for RGB-based policies in a new environment can require significant (e.g., months-long) human effort. In contrast, the pipeline we propose is data-driven and automated (with access only to policy training data and text descriptions of desired environmental changes). 

{\bf Anomaly detection and failure prediction.} Methods for \emph{failure prediction} seek to foresee failures as the robot is operating. Typical approaches include ones based on reachability analysis~\cite{akametalu2014reachability, hsu2023sim, hsu2023safety}, control barrier functions~\cite{ames2016control}, formal methods~\cite{alshiekh2018safe}, and learned predictors~\cite{farid2022failure, xie2022ask, gokmen2023asking, liu2024model}. A related line of work on \emph{anomaly detection} seeks to detect conditions that are far from nominal and may thus induce failures~\cite{richter2017safe, sinha2022system, salehi2021unified, sinha2024real,sindhwani2020unsupervised}. Our approach to predictive red teaming uses conformal prediction-based anomaly detection~\cite{vovk2005algorithmic, laxhammar2013online, luo2024sample, sinha2023closing}, which allows one to provide statistical assurances on the false positive rate of detection. Recently, conformal prediction has also been utilized in the context of robotics to provide statistical assurances on language-based planners, perception systems, and trajectory prediction systems~\cite{lindemann2023safe, dixit2023adaptive, ren2023robots, dixit2024perceive, lindemann2024formal}. All of the prior work mentioned above on failure prediction, anomaly detection, and conformal prediction develops methods that operate at \emph{runtime} in order to detect possible failures and take remedial measures. In contrast, we utilize anomaly detection to forecast performance in different environmental conditions by executing the detector on edited observations that reflect changes in these conditions. 

{\bf Generative image editing.} Prior work in robotics uses generative image editing~\cite{baldridge2024imagen,  betker2023improving, nichol2021glide, yu2023inpaint, ling2021editgan, zhu2020domain} for data augmentation~\cite{chen2023genaug, yu2023scaling, bharadhwaj2024roboagent, chen2024rovi, chen2024semantically}, generating sub-goals for image-conditioned policies~\cite{black2023zero, shah2023vint}, and runtime observation editing for visual generalization~\cite{hancock2024run}. In this work, we utilize a \emph{language-conditioned} image editing model (\texttt{Imagen 3}~\cite{baldridge2024imagen}) to generate image observations that reflect changes in various environmental factors (Fig.~\ref{fig:anchor}). By modifying real robot observations with targeted edits (e.g., ``change the background to red" or ``add a trash can to the scene"), we are able to generate synthetic observations with a high degree of realism. 

{\bf Off-policy evaluation.} The problem of predictive red teaming is related to \emph{off-policy evaluation} in reinforcement learning~\cite{precup2000eligibility, hallak2017consistent, hanna2017bootstrapping, farajtabar2018more}. The goal is to estimate the performance of a target policy using data collected by executing a different policy. This can be used for policy improvement, particularly in the offline reinforcement learning setting~\cite{levine2020offline}. Off-policy evaluation is similar to our goal of predictive teaming: both attempt to evaluate the performance of a policy without evaluating the policy on the robot. However, the two problems are also distinct: predictive red teaming attempts to predict the performance of a given policy in off-nominal conditions by executing the \emph{same} policy in nominal conditions. 

\section{Problem: Predictive Red Teaming}
\label{sec:problem formulation}


We formally introduce the problem of predictive red teaming: {\it exposing vulnerabilities of a given policy with respect to environmental factors such as lighting, visual distractors, and object locations, and predicting their impact on performance without performing any hardware evaluations in these off-nominal scenarios}.  

{\bf Nominal scenarios.} In each episode, the robot is deployed in a scenario $\xi$, which is defined as a partially observable Markov decision process (POMDP) initialized in a particular state. Let $\mathcal{D}_\text{nom}$ be a distribution over scenarios that captures nominal variations in all environmental factors (e.g., objects that the robot may encounter, lighting conditions, background colors, etc.) and tasks (via the reward function). We do not assume knowledge of $\mathcal{D}_\text{nom}$, except a dataset $S_\text{nom}$ of observations collected from nominal scenarios $\xi \sim \mathcal{D}_\text{nom}$. 

{\bf Inputs to the red team.} The \emph{red team} is provided a deterministic or stochastic policy $\pi$ that maps observations $o_t \in \mathcal{O}$ to actions $a_t \in \mathcal{A}$, along with the dataset $S_\text{nom}$ of nominal observations. Our focus in this paper will be on visuomotor policies trained via imitation learning; in this setting, $S_\text{nom}$ can consist of observations from the training dataset for $\pi$. We also assume access to a set $S_\text{val}$ of nominal observations that were held out when training $\pi$. The specific approach we present in this paper will only require nominal observations $S_\text{nom} \cup S_\text{val}$ collected at the \emph{start} of episodes. 
The red team is provided the ability to query $\pi$ on arbitrary observations, potentially with white-box access to internal representations of the policy. 

{\bf Goal: predictive red teaming.} The red team's goal is to expose vulnerabilities of $\pi$ with respect to various environmental factors $f \in F$ chosen by the red team.  These factors may be arbitrarily fine-grained, e.g., the introduction of a particular distractor or a specific change to the table color. Formally, let $\mathcal{D}_f$ be a distribution of scenarios where a factor $f$ has changed relative to the nominal distribution $\mathcal{D}_\text{nom}$. Let $R_\text{nom}^\pi$ be the expected reward of $\pi$ for scenarios $\xi \sim \mathcal{D}_\text{nom}$, and let $R_f^\pi$ be the expected reward for $\mathcal{D}_f$. For simplicity, we will assume henceforth that rewards are bounded in $[0,1]$. Knowing $R_\text{nom}^\pi$, we consider two problems: (1) \emph{rank} the factors $f \in F$ by performance degradation, and (2) predict the \emph{absolute} performance $R_f^\pi, \ \forall f \in F$. The former problem is important for targeted data collection, while the latter helps understand the envelope of acceptable performance.

\section{\redit: Predictive Red Teaming via Image Editing and Anomaly Detection}
\label{sec:approach}






We introduce \redit~({\bf Robo}tics {\bf A}uto-{\bf R}ed-{\bf T}eaming): a method for predictive red teaming using generative image editing and anomaly detection. We focus on visuomotor policies that rely primarily on RGB image observations. Our approach has two main steps, which are illustrated in Fig.~\ref{fig:anchor}. First, we use generative image editing tools to modify the nominal observations in $S_\text{val}$ (Sec.~\ref{sec:problem formulation}) to reflect changes in various factors of interest (e.g., background, lighting, distractor objects). For each factor, we then predict the performance degradation of the policy using anomaly detection. We describe each of these steps below. 

\subsection{Generative Image Editing}
\label{sec:image editing}

{\bf Selection of environmental factors.} The red team first selects a set $F$ of environmental factors that have the potential to degrade the performance of the given policy $\pi$. This set can be arbitrarily fine-grained in its contents (e.g., specific lighting conditions, distractor objects, background colors, etc.). The specific factors of interest will depend on the deployment needs of the policy and plausible environmental changes that the robot may encounter. 

{\bf Generating edited observations.} For each factor $f \in F$, we modify observations in the nominal set $S_\text{val}$ to reflect a change in $f$. We leverage state-of-the-art generative image editing tools, which have the capacity to take detailed language instructions as input in order to produce realistic and globally consistent edits. In this work, we specifically utilize the \texttt{Imagen~3} diffusion model~\cite{baldridge2024imagen}, which has been trained to perform language-prompted image editing tasks such as inpainting, outpainting, and colorization. 

\begin{figure}[t]
    \centering
    \includegraphics[width=\textwidth]{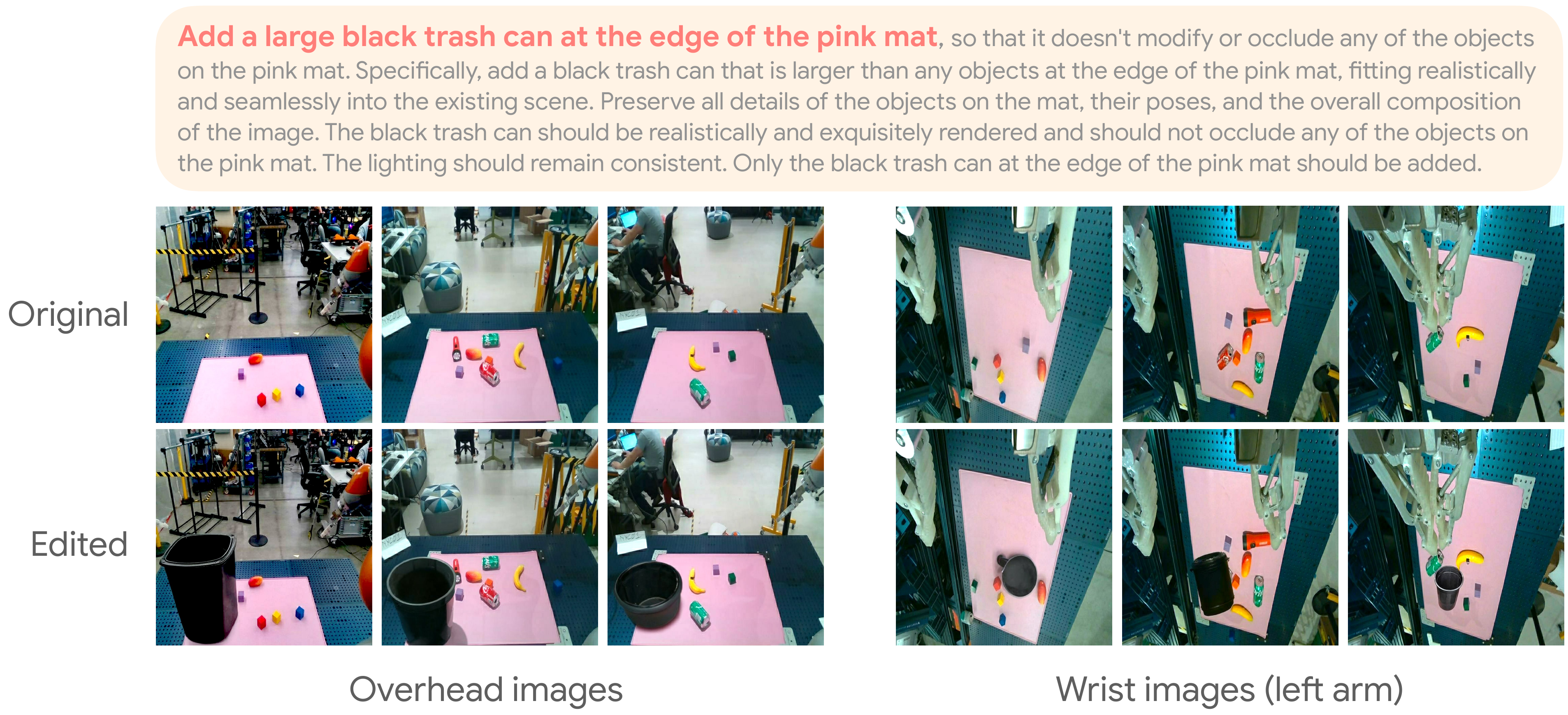}
    \caption{Examples of adding a novel object to the visual scene via generative image editing. Original (top) and edited (bottom) observations from both the robot's overhead and wrist cameras are shown. State-of-the-art generative image editing tools render the desired object in a realistic manner that maintains per-view global consistency in lighting, shadows, and overall composition.}
    \label{fig:distractor edit}
\end{figure}

As an example, consider an edit that adds a novel object to the scene. Fig.~\ref{fig:distractor edit} illustrates the prompt used for this edit, along with examples of the original and edited images. For robots with multiple cameras (e.g., a wrist camera in addition to an overhead camera), we edit each observation independently with the same prompt. Fig.~\ref{fig:distractor edit} shows the original and edited wrist camera images for the manipulator from Fig.~\ref{fig:anchor}. The image editing model is able to render the desired object in a realistic manner that maintains per-view global consistency in lighting, shadows, and overall composition of the scene (see Sec.~\ref{sec:future work} for a discussion of multi-view consistency). 

In addition to adding novel objects to the scene, state-of-the-art image editing models allow us to generate edits corresponding to various changes with a high degree of realism and precision, e.g., changing the color of the background, adding a human in the scene, and changing lighting conditions. Examples of these edits are illustrated in Fig.~\ref{fig:anchor}. Full prompts along with additional examples are provided in Appendix~\ref{app:image editing} and the \href{https://predictive-red-team.github.io/}{project website}.

\begin{figure}[h]
    \centering
    \includegraphics[width=\columnwidth]{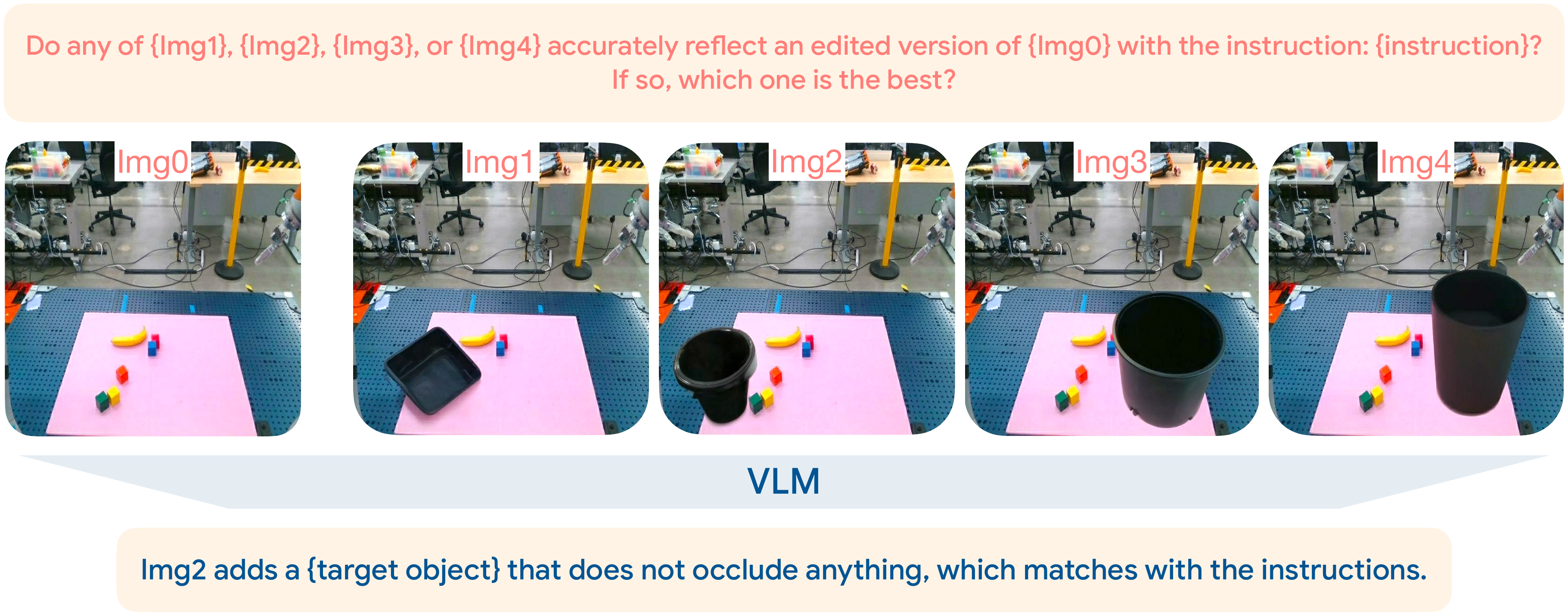}
    \caption{A vision-language model ensures that the edited image reflects the desired change.}
    \label{fig:vlm filter}
\vspace{-5pt}
\end{figure}

{\bf VLM critic.} Diffusion-based image editing models can generate multiple variations of edited images given the same input image and prompt. These variations often differ in terms of their quality and adherence to the prompt. In order to ensure that the edited observations accurately reflect the desired change in the environmental factor $f$, we generate a batch of four edited images per input, and utilize a vision-language model (VLM) as a \emph{critic}. As shown in Fig.~\ref{fig:vlm filter}, we prompt the VLM with the original and edited images, and ask it to judge if any of the options accurately reflect the desired change; if so, the VLM is tasked with choosing the best one (if not, we simply discard the observation from our set). The full prompt for the VLM --- which involves chain-of-thought reasoning --- is provided in Appendix~\ref{app:vlm filter}. We use the Gemini Pro 1.5 VLM~\cite{team2023gemini} for our experiments in Sec.~\ref{sec:experiments}.  

\subsection{Predicting Performance via Anomaly Detection}
\label{sec:anomaly detection}

At the end of the image editing process, the red team has a set $S_f$ of edited observations corresponding to each environmental factor $f \in F$. The second key component of \redit~(Fig.~\ref{fig:anchor}) uses $S_f$ to predict the performance degradation induced by each factor $f$. Our key idea is to utilize techniques from \emph{anomaly detection}: for each observation in $S_f$, we quantify how ``close" it is to nominal observations in $S_\text{nom}$ from the perspective of the policy $\pi$. If this distance is above a threshold computed using \emph{conformal prediction} \cite{vovk2005algorithmic}, the observation is flagged as an anomaly. The primary hypothesis is that one can define such a policy-specific anomaly detector that predicts performance degradation:
\begin{equation}
\label{eq:linear prediction}
R_f^\pi \approx 1 - \alpha_f^\pi,
\end{equation}
where $R_f^\pi$ is the expected reward under factor $f$ (Sec.~\ref{sec:problem formulation}) and $\alpha_f^\pi$ is the anomaly rate for $f$, i.e., the proportion of edited observations in $S_f$ flagged as anomalous according to a threshold chosen to ensure $R_\text{nom}^\pi \approx 1 - \alpha_\text{nom}^\pi$ (where $\alpha_\text{nom}^\pi$ is the proportion of nominal observations flagged as anomalous). 




{\bf Anomaly detection.} Next, we further describe how to compute the anomaly rate $\alpha_f^\pi$ for each factor $f$ using the edited observations $S_f$. In this work, we utilize policy embedding distances as a method for quantifying how far from nominal a given observation is. This choice is motivated by the prior success of embedding-based methods in anomaly detection (see, e.g., \cite{sinha2024real, luo2024online}) and the simplicity of implementation. Let $\phi_\pi(o)$ be a latent representation produced by the policy $\pi$ for a given observation $o$. For policies directly parameterized using a neural network, a common choice is to use the output of an intermediate layer of the network. In our experiments in Sec.~\ref{sec:experiments}, we utilize policies parameterized using diffusion models. In this setting, we utilize the context vector provided to the denoising process as our latent representation; see Appendix~\ref{app:policy} for more details. Using $\phi_\pi$, we can define a policy-specific anomaly score $s_\pi(o, S_\text{nom})$ that quantifies how far from nominal the observation $o$ is. A simple choice is to define $s_\pi$ as the nearest-neighbor cosine distance between the embedding $\phi_\pi(o)$ and the embeddings computed for the nominal observations in $S_\text{nom}$:
\begin{equation}
s_\pi(o, S_\text{nom}) := \min \ \left\{ 1 - \frac{\phi_\pi(o) \cdot \phi_\pi(o_i^\text{nom})}{\|\phi_\pi(o)\| \|\phi_\pi(o_i^\text{nom})\|} \ \middle| \ o_i \in S_\text{nom} \right\}. 
\end{equation}
A more general variant that we use in our experiments is to compute the mean of the $k$-nearest neighbor cosine distances. Intuitively, this anomaly score quantifies how dissimilar a given observation is compared to similar training observations from the perspective of the policy. 

For each factor $f \in F$, we compute the anomaly score for all edited observations $o \in S_f$. 
The anomaly rate for a factor $f$ is then defined as the proportion of observations flagged as anomalous according to a threshold $\tau$:
\begin{equation}
    \alpha_f^\pi := \frac{\left| \left\{ o \in S_f \mid s_\pi(o, S_\text{nom}) > \tau \right\} \right|}{|S_f|}.
\end{equation}

{\bf Anomaly threshold.} The anomaly threshold $\tau$ is chosen to ensure that $\alpha_\text{nom}^\pi$ (the anomaly rate for \emph{nominal} observations) predicts the nominal success rate $R_\text{nom}^\pi$ of the policy: $R_\text{nom}^\pi \approx 1 - \alpha_\text{nom}^\pi$. Given access to a validation set $S_\text{val}$ with $n_\text{val}$ nominal observations, one can simply choose $\tau$ such that the proportion of these flagged as anomalous is $1 - R_\text{nom}^\pi$. A more sophisticated approach uses conformal prediction~\cite{vovk2005algorithmic}: 
\begin{equation}
\label{eq:tau}
    \tau := \text{quantile}_{ \frac{\lceil (n_\text{val}+1)R_\text{nom}^\pi\rceil}{n_\text{val}}}(\{s_\pi(o, S_\text{nom}) \ | \ o \in S_\text{val} \}),
\end{equation}
which chooses $\tau$ as the $\lceil (n_\text{val}+1)R_\text{nom}^\pi\rceil/{n_\text{val}}$ empirical quantile of the set of anomaly scores for the validation set. This choice upper bounds the probability that \emph{unseen} nominal observations are flagged as anomalous to $1 - R_\text{nom}^\pi$~\cite{angelopoulos2021gentle}. 

\vspace{7pt}

We summarize the key steps of \redit~in Algorithm~\ref{alg:redit}.

\begin{algorithm}[h]
\caption{\redit: Robotics Auto Red Teaming}\label{alg:redit}
\begin{algorithmic}
\State {\bf Input: } Policy $\pi$ with nominal performance $R_\text{nom}^\pi$, nominal observations $S_\text{nom} \cup S_\text{val}$
\State Select environmental factors $F$
\State {\bf Conformal prediction:}
\State \hspace{2pt}  Compute anomaly scores for $S_\text{val}$ using $\pi$ embeddings:
\State \hspace{15pt} $\Lambda_\text{val} := \{s_\pi(o, S_\text{nom}) \ | \ o \in S_\text{val} \}$
\State \hspace{2pt}  Compute anomaly threshold $\tau$ using $\Lambda_\text{val}$ to bound the \\ \hspace{2pt} nominal anomaly rate to $\alpha_\text{nom}^\pi := 1 - R_\text{nom}^\pi$ \Comment{Eq.~\ref{eq:tau}}
\For{$f \in F$}
    \State Generate edited observations $S_f$ \Comment{Filtered with VLM}
\State Compute anomaly rate: 
\State \hspace{5pt} $\alpha_f^\pi := \left| \left\{ o \in S_f \mid s_\pi(o, S_\text{nom}) > \tau \right\} \right| / |S_f|$
\State Predict performance:
\State \hspace{5pt} $R_{f, \text{pred}}^\pi := 1 - \alpha_f^\pi$
\EndFor

\end{algorithmic}
\end{algorithm}
\section{Experiments}
\label{sec:experiments}

We evaluate our framework using 500+ hardware trials that vary twelve environmental factors (Fig.~\ref{fig:all factors}) for two visuomotor diffusion policies with significantly different architectures. These experiments seek to investigate the following questions: 
\begin{enumerate}
\item How well does \redit~identify vulnerabilities and predict policy performance when relevant environmental factors are varied? 
\item How effective is \redit~in enabling policy improvement via targeted data collection? 
\item How good of a proxy is anomaly detection for performance degradation in different environmental conditions?
\end{enumerate}

{\bf Hardware setup.} 
Fig.~\ref{fig:anchor} visualizes our hardware platform: a bimanual manipulator with two Kuka IIWA robotic arms (our experiments only utilize the left arm). We use a Weiss gripper to interact with the objects in the environment. For sensing, we use a dual camera setup with a fixed overhead camera and another camera mounted on the left wrist.

{\bf Training data.} We use a trajectory optimization-based motion planner to automatically collect a set of training data consisting of 3K+ demonstrations for grasping objects. These demonstrations are collected in nominal conditions, i.e., with fixed lighting, with a fixed pink background on a table, and an object set that consists of blocks, plush toys, small cans, and artificial fruits. Additional details on the data collection pipeline are provided in Appendix~\ref{app:policy}. We highlight that the chosen task (grasping) is relatively easy to learn, and hence makes the problem of red teaming \emph{more challenging}; trained policies demonstrate a nontrivial degree of generalization, but are also vulnerable in ways that are hard to intuit.

{\bf Policies.} We consider two policies that vary significantly in their overall architecture. The first policy, $\pi_\text{hyb}$, uses a hybrid policy architecture inspired by \cite{belkhale2023hydra}, which aims to combine the benefits of trajectory optimization for free-space planning with the reactive nature of closed-loop visuomotor diffusion policies~\cite{chi2023diffusion}. We achieve this by using two separate heads in a diffusion policy architecture: the first predicts a waypoint to be reached using trajectory optimization, and the second predicts a temporally dense sequence of actions. An additional head predicts which mode should be executed at any given time. All three heads are trained \emph{jointly} using a diffusion objective. The latent embedding (used by \redit~for anomaly detection) is a vector in $\mathbb{R}^{512}$ that encodes the robot's visual and proprioceptive observations, along with a keypoint selected by the robot's operator that serves as an instruction on which object to grasp.

We also separately train a vanilla diffusion policy~\cite{chi2023diffusion}, $\pi_\text{dfn}$, with a single head that outputs a trajectory sequence at every time-step (executed in a receding-horizon manner). The latent embedding for this policy is a vector in $\mathbb{R}^{515 \times 513}$. Details of the vision and instruction encoders, along with other implementation details, are provided in Appendix~\ref{app:policy}. Both policies achieve a success rate of approximately $65\%$ for nominal conditions, as measured by 30 trials (each) that vary objects, their locations, and the target object. 

{\bf Environmental factors.} We choose a set $F$ of twelve environmental factors that reflect common vulnerabilities of visuomotor policies trained via behavior cloning. These are shown in Fig.~\ref{fig:all factors}, and include: (1--3) three changes to the {\bf lighting} conditions (red, green, blue), (4--6) three changes to the color (red, green, blue) of the table {\bf background} on which objects are placed, (7--10) four {\bf distractor} objects (black and white trash can, laptop, candle) that partially occlude other objects, (11) a distractor in the form of a {\bf person} close to the table, and (12) a change to the {\bf height} of the table (which changes the location of objects relative to the overhead camera). In order to evaluate the predictions made by \redit, we execute both policies in 20+ episodes for each of the twelve factors; this allows us to estimate the corresponding success rates $R^{\pi_\text{hyb}}_f$ and $R^{\pi_\text{dfn}}_f, \forall f \in F$. The subsequent results thus include 500+ hardware trials. 

\subsection{How accurately does \redit~identify vulnerabilities and predict policy performance?}
\label{sec:redit experiments}

We first evaluate how well \redit~predicts the performance degradation induced by each of the twelve environmental factors for the different policies. We utilize two metrics to evaluate \redit, which correspond to the two versions of predictive red teaming described in Sec.~\ref{sec:problem formulation}: 
\begin{enumerate}
    \item {\bf Spearman rank correlation}~\cite{zar2005spearman}: this is a value $\rho \in [-1,1]$ which measures how accurately \redit \emph{ranks} the different factors by performance degradation. 
    \item {\bf Average prediction error}: measures how accurately \redit~predicts the \emph{absolute success rates} for the different factors by computing $\frac{1}{|F|}\sum_{f \in F} |R_{f}^\pi - R_{f, \text{pred}}^\pi|$. 
    \end{enumerate}
    
\begin{figure*}[t]
    \centering
    \includegraphics[width=0.8\textwidth]{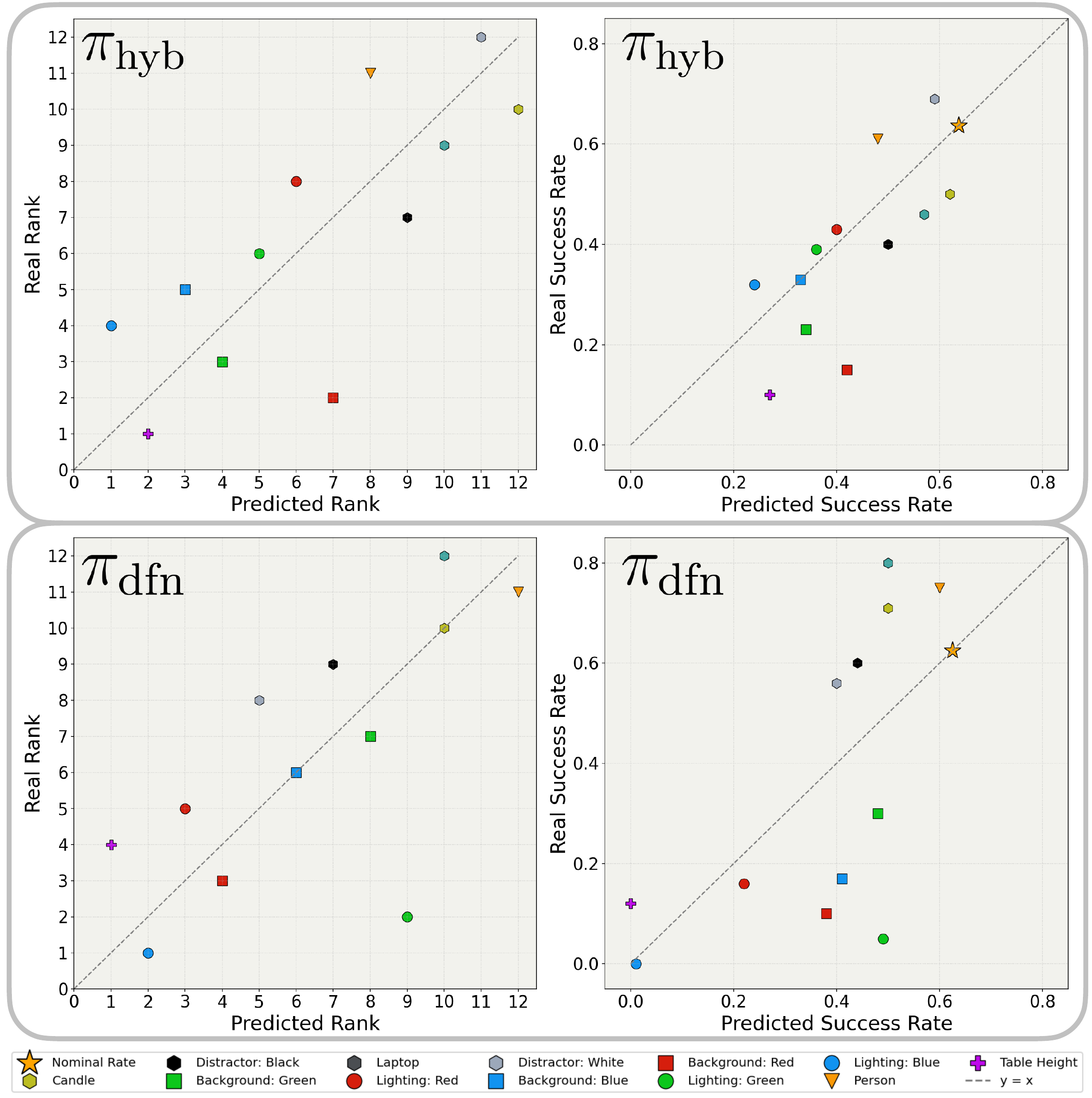}
    \caption{Evaluating predictions from \redit for  $\pi_\text{hyb}$ (which combines trajectory optimization with diffusion) in the top panel and $\pi_\text{hyb}$ (vanilla diffusion policy) in the bottom panel. Left: Comparison of true (estimated) rankings of different environmental factors by performance degradation with predictions made by \redit. Right: Comparison of true (estimated) success rates with predictions from \redit.}
    \label{fig:predictions hybrid}
    \vspace{-10pt}
\end{figure*}


{\bf Implementation.} In order to make predictions using \redit, we generate a set $S_f$ of 100 edited observations for each factor $f$ using first time-step observations from a held-out portion of training episodes. Examples of edits and complete prompts are provided in Fig.~\ref{fig:anchor} and Appendix~\ref{app:image editing}. We compute the resulting anomaly rates $\alpha_f^\pi$ using $S_f$ for each policy as described in Alg.~\ref{alg:redit}. 
We take the anomaly score $s_\pi(o, S_\text{nom})$ to be the mean of the $k$-nearest neighbor cosine distances (in the respective policy embedding space) to a set $S_\text{nom}$, which is chosen to be a subset of first-time-step observations from the training episodes.

{\bf Results.} Fig.~\ref{fig:predictions hybrid} evaluates predictions made by \redit~for $\pi_\text{hyb}$ (top row) and $\pi_\text{dfn}$ (bottom row). Fig.~\ref{fig:predictions hybrid}-left compares the rankings of different environmental factors $f \in F$ predicted by \redit with the true rankings as measured by the 20+ hardware evaluations for each factor (a lower rank corresponds to a lower success rate). Fig.~\ref{fig:predictions hybrid}-right compares the absolute success rates predicted by \redit~with the true (estimated) success rates. As the figure illustrates, the predictions for both rankings and absolute performance are strongly correlated with the true rankings and success rates. For example, \redit~successfully identifies that $\pi_\text{hyb}$ is relatively robust to object or person distractors, moderately impacted by changes in the background and lighting, and strongly impacted by changing the height of the table. \redit~also successfully identifies that $\pi_\text{dfn}$ is more vulnerable than $\pi_\text{hyb}$ to certain environmental factors such as blue lighting and a change in the table height.  

Table~\ref{tab:predictions} quantitatively evaluates the predictions made by \redit~for both policies. The Spearman $\rho$ indicates a strong correlation between the predicted and actual rankings of factors, while the average prediction error is under $0.19$ for both policies (which is roughly in the range of noise when estimating success rates from $\sim 20$ trials). 

\begin{table}[h]
\centering
\resizebox{0.5\linewidth}{!}{%
\begin{tabular}{@{}lll@{}}
\toprule
           \redit        & $\pi_\text{hyb}$ &    $\pi_\text{dfn}$  \\ \midrule
{\bf Spearman}  $\rho \in [-1,1]$           & ~0.8  ($\uparrow$)     & ~0.7 ($\uparrow$)  \\
{\bf Av. prediction error} $ \in [0,1]$    & 0.10  ($\downarrow$)  & 0.19 ($\downarrow$) \\ \bottomrule
\end{tabular}%
}
\caption{Quantitative evaluation of success rates predicted by \redit compared with real success rates.}
\label{tab:predictions}
\vspace{-5pt}
\end{table}

{\bf Ablations.} For the results above, we use $k=5, |S_\text{nom}| = 3000$ for $\pi_\text{hyb}$ and $k=10, |S_\text{nom}| = 500$ for $\pi_\text{dfn}$.  We provide results from ablating the values $k$ and $|S_\text{nom}|$ in Appendix~\ref{app:ablations}. Generally, we find that predictions for $\pi_\text{hyb}$ remain accurate when varying $|S_\text{nom}|$ with small $k$, while predictions for $\pi_\text{dfn}$ (which has a significantly higher dimensional embedding space) benefit from either having a smaller value of $|S_\text{nom}|$ or larger values of $k$. 

\subsection{How effective is \redit~in enabling policy improvement via targeted data collection?}
\label{subsec:targeted-data-collection}


Our second set of experiments seeks to evaluate how well predictions from \redit~enable \emph{policy improvement} via targeted data collection. To this end, we collect additional demonstration data for $\pi_\text{hyb}$ with the three environmental factors that \redit~predicts the highest performance degradation for: blue lighting, change in the table height, and blue table background.
We collect around 1 hour of training data ($\approx 100$ trajectories) under each of these off-nominal settings. We then co-finetune our initial learned policy $\pi_\text{hyb}$ on the older data combined with the new small amount of off-nominal data. During co-finetuning, we ensure that each mini-batch consists of $80\%$ of the original data mixture and $20\%$ from the new off-nominal data. We co-finetune the policy with a reduced learning rate ($5e-6$) for a total of 20K steps.  

\begin{figure}[h]
    \centering
    \includegraphics[width=0.7\columnwidth]{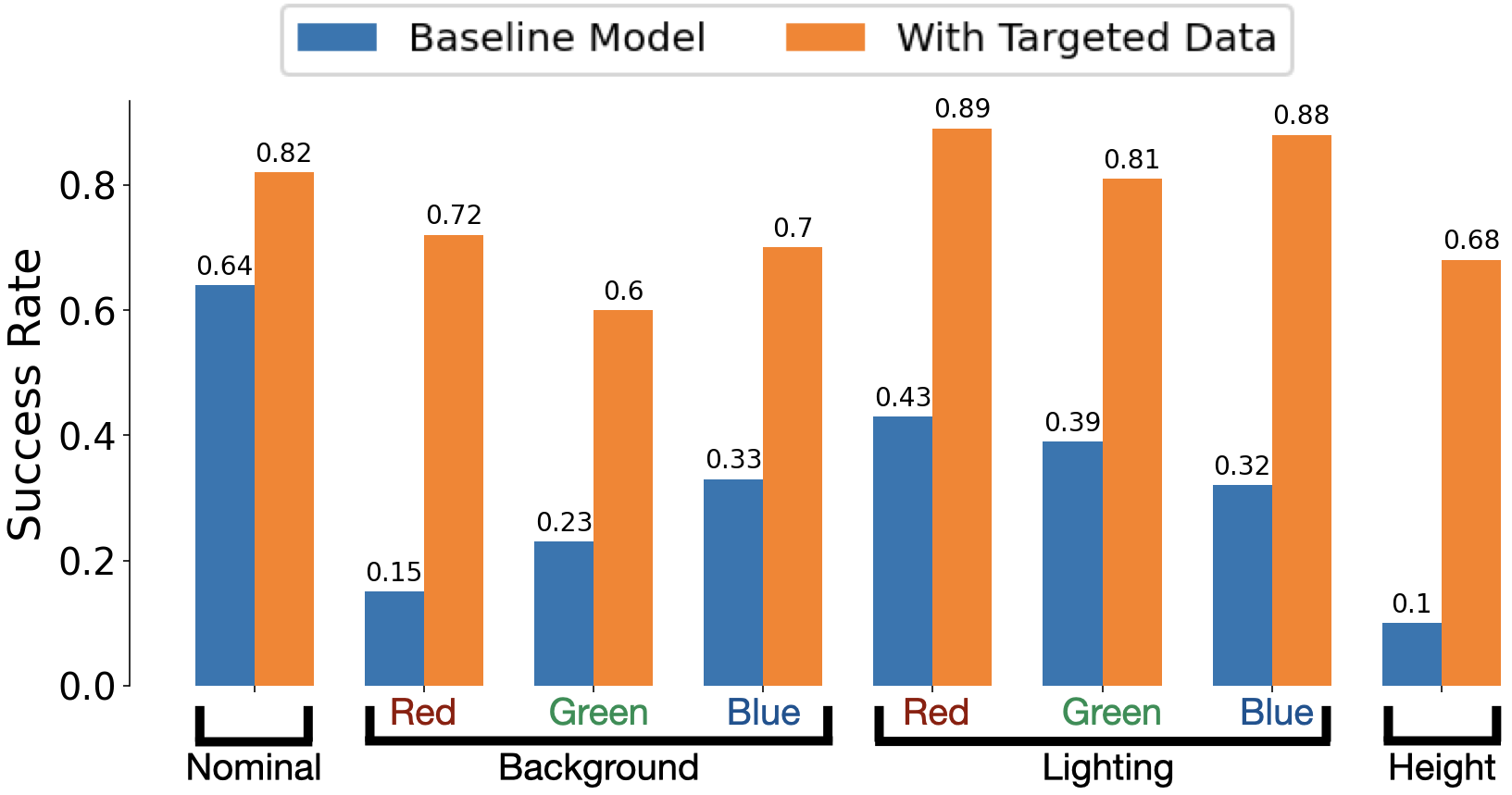}
    \caption{Fine-tuning with data collected under conditions predicted to be adverse shows cross-domain generalization and boosts baseline performance by 2--7x.}
    \label{fig:finetuning}
\vspace{-10pt}
\end{figure}

The fine-tuned policy $\pi_\text{hyb}^\text{ft}$ is evaluated in nominal conditions, the three conditions for which we collected data, and also the other background and lighting conditions. Videos of $\pi_\text{hyb}^\text{ft}$ are available on the \href{https://predictive-red-team.github.io/}{project website}. Fig.~\ref{fig:finetuning} shows the results. We observe improved performance in nominal conditions and a 2--7x improvement in off-nominal conditions under which training data was collected. Interestingly, the targeted data collection also yields \emph{cross-domain generalization}: the performance of the policy is improved by 2--5x even for background and lighting conditions where we did not collect additional data. This highlights the benefits of targeting data collection towards adverse scenarios via predictive red teaming. 

\subsection{How accurately does anomaly detection predict performance degradation?} 
\label{sec:anomaly-to-real}

Our final set of experiments seeks to evaluate the anomaly detection component of \redit~in isolation from the image editing pipeline. To this end, instead of executing the embedding-based anomaly detector on the set $S_f$ of \emph{edited} observations (as described in Algorithm~\ref{alg:redit}), we execute the detector on the set $S_f^\text{real}$ composed of \emph{real} robot observations collected from the first time step of the 20+ episodes where the factor $f$ is varied. We then compute the corresponding anomaly rates $\alpha_{f, \text{real}}^\pi$ ($\forall f \in F$). Predicted success rates for each factor are computed as  $R_{f, \text{anom}}^\pi := 1 - \alpha_{f, \text{real}}^\pi$ and compared with the (estimated) true success rates. Additional implementation details are provided in Appendix~\ref{app:anomaly-to-real}.

Table~\ref{tab:anomaly-to-real} presents the Spearman $\rho$ and average prediction errors. Appendix~\ref{app:anomaly-to-real} presents figures analogous to Fig.~\ref{fig:predictions hybrid}. While we observe high values of $\rho$ and low values of the prediction error for both policies, we note that the predictions $R_{f, \text{anom}}^\pi$ are made using $5\times$ fewer observations than predictions from the full \redit~pipeline ($\sim 20$ real observations vs. $100$ edited observations), thus making them significantly more susceptible to noise. 

\begin{table}[t]
\centering
\resizebox{0.5\linewidth}{!}{%
\begin{tabular}{@{}lll@{}}
\toprule
           Anomaly detector                 & $\pi_\text{hyb}$              &    $\pi_\text{dfn}$  \\ \midrule
{\bf Spearman}  $\rho \in [-1,1]$           & ~0.6  ($\uparrow$)   & ~0.8 ($\uparrow$)  \\
{\bf Av. prediction error} $ \in [0,1]$    & 0.20 ($\downarrow$)           & 0.21 ($\downarrow$) \\ \bottomrule
\end{tabular}%
}
\caption{Evaluating predictions of success rates made from anomaly rates computed using real observations.}
\label{tab:anomaly-to-real}
\vspace{-5pt}
\end{table}


\section{Conclusions}
\label{sec:conclusions}

{\bf Summary.} We have introduced the problem of predictive red teaming: given access to observations from nominal scenarios, discovering vulnerabilities of a policy with respect to unseen changes in environmental factors and predicting the resulting performance degradation. Our approach to predictive red teaming --- Robotics Auto Red Teaming (\redit) --- modifies nominal observations via generative image editing to reflect changes in environmental factors of interest, and then uses a policy embedding-based anomaly detector to predict performance degradation. Experiments across 500+ trials for visuomotor diffusion policies demonstrate \redit's ability to (i) identify environmental factors that significantly impact performance, (ii) predict the relative and absolute impact of factors, and (iii) enable policy improvement via targeted data collection.   

\subsection{Limitations and Future Work}
\label{sec:future work}

We discuss limitations of our approach and promising future directions that may address them. 

{\bf Edit-to-real gap.} While state-of-the-art image editing tools are capable of producing realistic edits (especially with careful prompting), there are still gaps in realism for certain environmental factors. For example, edits that reflect lighting changes (Fig.~\ref{fig:anchor}) do not modify the shadows of objects as real lighting changes do. We expect that our method will benefit from the significant investments in improving image editing models for commercial applications. 
Beyond single-view realism, a more challenging limitation is ensuring \emph{multi-view} consistency. As seen in Fig.~\ref{fig:distractor edit}, edited observations from the overhead and wrist cameras do not represent a consistent geometry for the new object. One exciting possibility is to utilize recent \emph{3D scene editing} tools based on Gaussian Splatting~\cite{chen2024proedit, bao20243d} that allow for edits with multi-view consistency. Scene editing may also allow us to go beyond RGB observations and edit depth channels. 

{\bf Anomaly-to-failure gap.} Our approach utilizes the anomaly rate as a predictor for performance degradation. However, as seen in Sec.~\ref{sec:anomaly-to-real}, these predictions are not perfectly accurate. One avenue for future work is to perform edits on observations from multiple time-steps within each episode, and to compute anomaly rates based on these sequences (rather than only utilizing the first time-step from episodes, as we currently do). We are also interested in exploring other methods from the vast literature on anomaly detection to identify techniques that may serve as better predictors of performance degradation (see, e.g., \cite{xuuncertainty} for a recent empirical study).

{\bf Hidden environmental factors.} An important limitation of \redit~is that it requires changes in environmental factors to be reflected in visual observations of the robot. As such, \redit~will not identify vulnerabilities with changes that are completely hidden (e.g., changing the mass of objects without changing their visual appearance). In such cases, predictive red teaming via simulation is a promising avenue, but requires bridging the sim-to-real gap, which is typically very significant for RGB observations and may be even more pronounced when simulating unseen off-nominal scenarios. 

{\bf Multi-round predictive red teaming.} \redit~currently chooses a single set $F$ of environmental factors at the beginning of predictive red teaming. A more sophisticated approach could \emph{iteratively} explore the space of environmental factors: choose an initial set $F$, make predictions for these, and expand the set iteratively to include factors that are similar to ones that were predicted to yield poor performance. 


\vspace{7pt}

As we seek to deploy robots in challenging real-world applications, it is essential that we develop scalable methods for predicting the limits of their performance. We hope that formalizing the problem of predictive red teaming and providing a baseline in the form of \redit~spurs further research along this underexplored direction.


\section*{Acknowledgments}
We are grateful to Shixin Luo and Suraj Kothawade for extremely helpful pointers on the image editing pipeline. We thank Alex Irpan for insightful feedback on an early version of the paper. We are grateful to Dave Orr, Anca Dragan, and Vincent Vanhoucke for continuous guidance on safety-related topics; and Frankie Garcia and Dawn Bloxwich on responsible development. 


\balance
\bibliographystyle{unsrtnat}
\bibliography{references}

\begin{thebibliography}{77}
\providecommand{\natexlab}[1]{#1}
\providecommand{\url}[1]{\texttt{#1}}
\expandafter\ifx\csname urlstyle\endcsname\relax
  \providecommand{\doi}[1]{doi: #1}\else
  \providecommand{\doi}{doi: \begingroup \urlstyle{rm}\Url}\fi

\bibitem[Chi et~al.(2023)Chi, Xu, Feng, Cousineau, Du, Burchfiel, Tedrake, and
  Song]{chi2023diffusion}
Cheng Chi, Zhenjia Xu, Siyuan Feng, Eric Cousineau, Yilun Du, Benjamin
  Burchfiel, Russ Tedrake, and Shuran Song.
\newblock Diffusion policy: Visuomotor policy learning via action diffusion.
\newblock \emph{The International Journal of Robotics Research}, 2023.

\bibitem[Baldridge et~al.(2024)Baldridge, Bauer, Bhutani, Brichtova, Bunner,
  Chan, Chen, Dieleman, Du, Eaton-Rosen, et~al.]{baldridge2024imagen}
Jason Baldridge, Jakob Bauer, Mukul Bhutani, Nicole Brichtova, Andrew Bunner,
  Kelvin Chan, Yichang Chen, Sander Dieleman, Yuqing Du, Zach Eaton-Rosen,
  et~al.
\newblock Imagen 3.
\newblock \emph{arXiv preprint arXiv:2408.07009}, 2024.

\bibitem[Esser et~al.(2024)Esser, Kulal, Blattmann, Entezari, M{\"u}ller,
  Saini, Levi, Lorenz, Sauer, Boesel, et~al.]{esser2024scaling}
Patrick Esser, Sumith Kulal, Andreas Blattmann, Rahim Entezari, Jonas
  M{\"u}ller, Harry Saini, Yam Levi, Dominik Lorenz, Axel Sauer, Frederic
  Boesel, et~al.
\newblock Scaling rectified flow transformers for high-resolution image
  synthesis.
\newblock In \emph{Proceedings of the International Conference on Machine
  Learning}, 2024.

\bibitem[Zhang et~al.(2023)Zhang, Rao, and Agrawala]{controlnet}
Lvmin Zhang, Anyi Rao, and Maneesh Agrawala.
\newblock Adding conditional control to text-to-image diffusion models.
\newblock In \emph{Proceedings of the IEEE/CVF International Conference on
  Computer Vision}, pages 3836--3847, 2023.

\bibitem[Saharia et~al.(2022)Saharia, Chan, Saxena, Li, Whang, Denton,
  Ghasemipour, Gontijo~Lopes, Karagol~Ayan, Salimans,
  et~al.]{saharia2022photorealistic}
Chitwan Saharia, William Chan, Saurabh Saxena, Lala Li, Jay Whang, Emily~L
  Denton, Kamyar Ghasemipour, Raphael Gontijo~Lopes, Burcu Karagol~Ayan, Tim
  Salimans, et~al.
\newblock Photorealistic text-to-image diffusion models with deep language
  understanding.
\newblock \emph{Advances in neural information processing systems},
  35:\penalty0 36479--36494, 2022.

\bibitem[Vovk et~al.(2005)Vovk, Gammerman, and Shafer]{vovk2005algorithmic}
Vladimir Vovk, Alexander Gammerman, and Glenn Shafer.
\newblock \emph{Algorithmic Learning in a Random World}, volume~29.
\newblock Springer, 2005.

\bibitem[Zenko(2015)]{zenko2015red}
Micah Zenko.
\newblock \emph{Red Team: How To Succeed By Thinking Like The Enemy}.
\newblock Basic Books, 2015.

\bibitem[Achiam et~al.(2023)Achiam, Adler, Agarwal, Ahmad, Akkaya, Aleman,
  Almeida, Altenschmidt, Altman, Anadkat, et~al.]{achiam2023gpt}
Josh Achiam, Steven Adler, Sandhini Agarwal, Lama Ahmad, Ilge Akkaya,
  Florencia~Leoni Aleman, Diogo Almeida, Janko Altenschmidt, Sam Altman,
  Shyamal Anadkat, et~al.
\newblock {GPT}-4 technical report.
\newblock \emph{arXiv preprint arXiv:2303.08774}, 2023.

\bibitem[Team et~al.(2023)Team, Anil, Borgeaud, Alayrac, Yu, Soricut,
  Schalkwyk, Dai, Hauth, Millican, et~al.]{team2023gemini}
Gemini Team, Rohan Anil, Sebastian Borgeaud, Jean-Baptiste Alayrac, Jiahui Yu,
  Radu Soricut, Johan Schalkwyk, Andrew~M Dai, Anja Hauth, Katie Millican,
  et~al.
\newblock Gemini: a family of highly capable multimodal models.
\newblock \emph{arXiv preprint arXiv:2312.11805}, 2023.

\bibitem[Ganguli et~al.(2022)Ganguli, Lovitt, Kernion, Askell, Bai, Kadavath,
  Mann, Perez, Schiefer, Ndousse, et~al.]{ganguli2022red}
Deep Ganguli, Liane Lovitt, Jackson Kernion, Amanda Askell, Yuntao Bai, Saurav
  Kadavath, Ben Mann, Ethan Perez, Nicholas Schiefer, Kamal Ndousse, et~al.
\newblock Red teaming language models to reduce harms: Methods, scaling
  behaviors, and lessons learned.
\newblock \emph{arXiv preprint arXiv:2209.07858}, 2022.

\bibitem[Bai et~al.(2022)Bai, Jones, Ndousse, Askell, Chen, DasSarma, Drain,
  Fort, Ganguli, Henighan, et~al.]{bai2022training}
Yuntao Bai, Andy Jones, Kamal Ndousse, Amanda Askell, Anna Chen, Nova DasSarma,
  Dawn Drain, Stanislav Fort, Deep Ganguli, Tom Henighan, et~al.
\newblock Training a helpful and harmless assistant with reinforcement learning
  from human feedback.
\newblock \emph{arXiv preprint arXiv:2204.05862}, 2022.

\bibitem[Wei et~al.(2024)Wei, Haghtalab, and Steinhardt]{wei2024jailbroken}
Alexander Wei, Nika Haghtalab, and Jacob Steinhardt.
\newblock Jailbroken: How does {LLM} safety training fail?
\newblock \emph{Advances in Neural Information Processing Systems}, 36, 2024.

\bibitem[Perez et~al.(2022)Perez, Huang, Song, Cai, Ring, Aslanides, Glaese,
  McAleese, and Irving]{perez2022red}
Ethan Perez, Saffron Huang, Francis Song, Trevor Cai, Roman Ring, John
  Aslanides, Amelia Glaese, Nat McAleese, and Geoffrey Irving.
\newblock Red teaming language models with language models.
\newblock \emph{arXiv preprint arXiv:2202.03286}, 2022.

\bibitem[Tong et~al.(2024)Tong, Jones, and Steinhardt]{tong2024mass}
Shengbang Tong, Erik Jones, and Jacob Steinhardt.
\newblock Mass-producing failures of multimodal systems with language models.
\newblock \emph{Advances in Neural Information Processing Systems}, 36, 2024.

\bibitem[Zou et~al.(2023)Zou, Wang, Carlini, Nasr, Kolter, and
  Fredrikson]{zou2023universal}
Andy Zou, Zifan Wang, Nicholas Carlini, Milad Nasr, J~Zico Kolter, and Matt
  Fredrikson.
\newblock Universal and transferable adversarial attacks on aligned language
  models.
\newblock \emph{arXiv preprint arXiv:2307.15043}, 2023.

\bibitem[Chao et~al.(2023)Chao, Robey, Dobriban, Hassani, Pappas, and
  Wong]{chao2023jailbreaking}
Patrick Chao, Alexander Robey, Edgar Dobriban, Hamed Hassani, George~J Pappas,
  and Eric Wong.
\newblock Jailbreaking black box large language models in twenty queries.
\newblock \emph{arXiv preprint arXiv:2310.08419}, 2023.

\bibitem[Liu et~al.(2023)Liu, Xu, Chen, and Xiao]{liu2023autodan}
Xiaogeng Liu, Nan Xu, Muhao Chen, and Chaowei Xiao.
\newblock Auto{DAN}: Generating stealthy jailbreak prompts on aligned large
  language models.
\newblock \emph{arXiv preprint arXiv:2310.04451}, 2023.

\bibitem[Mehrotra et~al.(2023)Mehrotra, Zampetakis, Kassianik, Nelson,
  Anderson, Singer, and Karbasi]{mehrotra2023tree}
Anay Mehrotra, Manolis Zampetakis, Paul Kassianik, Blaine Nelson, Hyrum
  Anderson, Yaron Singer, and Amin Karbasi.
\newblock Tree of attacks: Jailbreaking black-box {LLM}s automatically.
\newblock \emph{arXiv preprint arXiv:2312.02119}, 2023.

\bibitem[Li et~al.(2024)Li, Li, Yin, Ahmed, Liu, and Liu]{li2024red}
Mukai Li, Lei Li, Yuwei Yin, Masood Ahmed, Zhenguang Liu, and Qi~Liu.
\newblock Red teaming visual language models.
\newblock \emph{arXiv preprint arXiv:2401.12915}, 2024.

\bibitem[Liu et~al.(2024{\natexlab{a}})Liu, Cai, Zhang, Yuan, and
  Wang]{liu2024arondight}
Yi~Liu, Chengjun Cai, Xiaoli Zhang, Xingliang Yuan, and Cong Wang.
\newblock Arondight: Red teaming large vision language models with
  auto-generated multi-modal jailbreak prompts.
\newblock In \emph{Proceedings of the 32nd ACM International Conference on
  Multimedia}, pages 3578--3586, 2024{\natexlab{a}}.

\bibitem[Rando et~al.(2022)Rando, Paleka, Lindner, Heim, and
  Tram{\`e}r]{rando2022red}
Javier Rando, Daniel Paleka, David Lindner, Lennart Heim, and Florian
  Tram{\`e}r.
\newblock Red-teaming the stable diffusion safety filter.
\newblock \emph{arXiv preprint arXiv:2210.04610}, 2022.

\bibitem[Gandikota et~al.(2023)Gandikota, Materzynska, Fiotto-Kaufman, and
  Bau]{gandikota2023erasing}
Rohit Gandikota, Joanna Materzynska, Jaden Fiotto-Kaufman, and David Bau.
\newblock Erasing concepts from diffusion models.
\newblock In \emph{Proceedings of the IEEE/CVF International Conference on
  Computer Vision}, pages 2426--2436, 2023.

\bibitem[Karnik et~al.(2024)Karnik, Hong, Abhangi, Lin, Wang, and
  Agrawal]{karnik2024embodied}
Sathwik Karnik, Zhang-Wei Hong, Nishant Abhangi, Yen-Chen Lin, Tsun-Hsuan Wang,
  and Pulkit Agrawal.
\newblock Embodied red teaming for auditing robotic foundation models.
\newblock \emph{arXiv preprint arXiv:2411.18676}, 2024.

\bibitem[Robey et~al.(2024)Robey, Ravichandran, Kumar, Hassani, and
  Pappas]{robey2024jailbreaking}
Alexander Robey, Zachary Ravichandran, Vijay Kumar, Hamed Hassani, and George~J
  Pappas.
\newblock Jailbreaking {LLM}-controlled robots.
\newblock \emph{arXiv preprint arXiv:2410.13691}, 2024.

\bibitem[Pumacay et~al.(2024)Pumacay, Singh, Duan, Krishna, Thomason, and
  Fox]{pumacay2024colosseum}
Wilbert Pumacay, Ishika Singh, Jiafei Duan, Ranjay Krishna, Jesse Thomason, and
  Dieter Fox.
\newblock {THE COLOSSEUM}: A benchmark for evaluating generalization for
  robotic manipulation.
\newblock 2024.

\bibitem[Akametalu et~al.(2014)Akametalu, Fisac, Gillula, Kaynama, Zeilinger,
  and Tomlin]{akametalu2014reachability}
Anayo~K Akametalu, Jaime~F Fisac, Jeremy~H Gillula, Shahab Kaynama, Melanie~N
  Zeilinger, and Claire~J Tomlin.
\newblock Reachability-based safe learning with gaussian processes.
\newblock In \emph{Proceedings of the 53rd IEEE Conference on Decision and
  Control}, pages 1424--1431, 2014.

\bibitem[Hsu et~al.(2023{\natexlab{a}})Hsu, Ren, Nguyen, Majumdar, and
  Fisac]{hsu2023sim}
Kai-Chieh Hsu, Allen~Z Ren, Duy~P Nguyen, Anirudha Majumdar, and Jaime~F Fisac.
\newblock Sim-to-lab-to-real: Safe reinforcement learning with shielding and
  generalization guarantees.
\newblock \emph{Artificial Intelligence}, 314:\penalty0 103811,
  2023{\natexlab{a}}.

\bibitem[Hsu et~al.(2023{\natexlab{b}})Hsu, Hu, and Fisac]{hsu2023safety}
Kai-Chieh Hsu, Haimin Hu, and Jaime~F Fisac.
\newblock The safety filter: A unified view of safety-critical control in
  autonomous systems.
\newblock \emph{Annual Review of Control, Robotics, and Autonomous Systems}, 7,
  2023{\natexlab{b}}.

\bibitem[Ames et~al.(2016)Ames, Xu, Grizzle, and Tabuada]{ames2016control}
Aaron~D Ames, Xiangru Xu, Jessy~W Grizzle, and Paulo Tabuada.
\newblock Control barrier function based quadratic programs for safety critical
  systems.
\newblock \emph{IEEE Transactions on Automatic Control}, 62\penalty0
  (8):\penalty0 3861--3876, 2016.

\bibitem[Alshiekh et~al.(2018)Alshiekh, Bloem, Ehlers, K{\"o}nighofer, Niekum,
  and Topcu]{alshiekh2018safe}
Mohammed Alshiekh, Roderick Bloem, R{\"u}diger Ehlers, Bettina K{\"o}nighofer,
  Scott Niekum, and Ufuk Topcu.
\newblock Safe reinforcement learning via shielding.
\newblock In \emph{Proceedings of the AAAI Conference on Artificial
  Intelligence}, volume~32, 2018.

\bibitem[Farid et~al.(2022)Farid, Snyder, Ren, and Majumdar]{farid2022failure}
Alec Farid, David Snyder, Allen~Z. Ren, and Anirudha Majumdar.
\newblock Failure prediction with statistical guarantees for vision-based robot
  control.
\newblock In \emph{Proceedings of Robotics: Science and Systems (RSS)}, 2022.

\bibitem[Xie et~al.(2022)Xie, Tajwar, Sharma, and Finn]{xie2022ask}
Annie Xie, Fahim Tajwar, Archit Sharma, and Chelsea Finn.
\newblock When to ask for help: Proactive interventions in autonomous
  reinforcement learning.
\newblock \emph{Advances in Neural Information Processing Systems},
  35:\penalty0 16918--16930, 2022.

\bibitem[Gokmen et~al.(2023)Gokmen, Ho, and Khansari]{gokmen2023asking}
Cem Gokmen, Daniel Ho, and Mohi Khansari.
\newblock Asking for help: Failure prediction in behavioral cloning through
  value approximation.
\newblock In \emph{Proceedings of the IEEE International Conference on Robotics
  and Automation (ICRA)}, pages 5821--5828, 2023.

\bibitem[Liu et~al.(2024{\natexlab{b}})Liu, Dass, Mart{\'\i}n-Mart{\'\i}n, and
  Zhu]{liu2024model}
Huihan Liu, Shivin Dass, Roberto Mart{\'\i}n-Mart{\'\i}n, and Yuke Zhu.
\newblock Model-based runtime monitoring with interactive imitation learning.
\newblock In \emph{Proceedings of the IEEE International Conference on Robotics
  and Automation (ICRA)}, pages 4154--4161, 2024{\natexlab{b}}.

\bibitem[Richter and Roy(2017)]{richter2017safe}
Charles Richter and Nicholas Roy.
\newblock Safe visual navigation via deep learning and novelty detection.
\newblock In \emph{Proceedings of Robotics: Science and Systems (RSS)}, 2017.

\bibitem[Sinha et~al.(2022)Sinha, Sharma, Banerjee, Lew, Luo, Richards, Sun,
  Schmerling, and Pavone]{sinha2022system}
Rohan Sinha, Apoorva Sharma, Somrita Banerjee, Thomas Lew, Rachel Luo,
  Spencer~M Richards, Yixiao Sun, Edward Schmerling, and Marco Pavone.
\newblock A system-level view on out-of-distribution data in robotics.
\newblock \emph{arXiv preprint arXiv:2212.14020}, 2022.

\bibitem[Salehi et~al.(2021)Salehi, Mirzaei, Hendrycks, Li, Rohban, and
  Sabokrou]{salehi2021unified}
Mohammadreza Salehi, Hossein Mirzaei, Dan Hendrycks, Yixuan Li,
  Mohammad~Hossein Rohban, and Mohammad Sabokrou.
\newblock A unified survey on anomaly, novelty, open-set, and
  out-of-distribution detection: Solutions and future challenges.
\newblock \emph{arXiv preprint arXiv:2110.14051}, 2021.

\bibitem[Sinha et~al.(2024)Sinha, Elhafsi, Agia, Foutter, Schmerling, and
  Pavone]{sinha2024real}
Rohan Sinha, Amine Elhafsi, Christopher Agia, Matthew Foutter, Edward
  Schmerling, and Marco Pavone.
\newblock Real-time anomaly detection and reactive planning with large language
  models.
\newblock \emph{arXiv preprint arXiv:2407.08735}, 2024.

\bibitem[Sindhwani et~al.(2020)Sindhwani, Sidahmed, Choromanski, and
  Jones]{sindhwani2020unsupervised}
Vikas Sindhwani, Hakim Sidahmed, Krzysztof Choromanski, and Brandon Jones.
\newblock Unsupervised anomaly detection for self-flying delivery drones.
\newblock In \emph{2020 IEEE international conference on robotics and
  automation (ICRA)}, pages 186--192. IEEE, 2020.

\bibitem[Laxhammar and Falkman(2013)]{laxhammar2013online}
Rikard Laxhammar and G{\"o}ran Falkman.
\newblock Online learning and sequential anomaly detection in trajectories.
\newblock \emph{IEEE Transactions on Pattern Analysis and Machine
  Intelligence}, 36\penalty0 (6):\penalty0 1158--1173, 2013.

\bibitem[Luo et~al.(2024{\natexlab{a}})Luo, Zhao, Kuck, Ivanovic, Savarese,
  Schmerling, and Pavone]{luo2024sample}
Rachel Luo, Shengjia Zhao, Jonathan Kuck, Boris Ivanovic, Silvio Savarese,
  Edward Schmerling, and Marco Pavone.
\newblock Sample-efficient safety assurances using conformal prediction.
\newblock \emph{The International Journal of Robotics Research}, 43\penalty0
  (9):\penalty0 1409--1424, 2024{\natexlab{a}}.

\bibitem[Sinha et~al.(2023)Sinha, Schmerling, and Pavone]{sinha2023closing}
Rohan Sinha, Edward Schmerling, and Marco Pavone.
\newblock Closing the loop on runtime monitors with fallback-safe {MPC}.
\newblock In \emph{2023 62nd IEEE Conference on Decision and Control (CDC)},
  pages 6533--6540. IEEE, 2023.

\bibitem[Lindemann et~al.(2023)Lindemann, Cleaveland, Shim, and
  Pappas]{lindemann2023safe}
Lars Lindemann, Matthew Cleaveland, Gihyun Shim, and George~J Pappas.
\newblock Safe planning in dynamic environments using conformal prediction.
\newblock \emph{IEEE Robotics and Automation Letters}, 2023.

\bibitem[Dixit et~al.(2023)Dixit, Lindemann, Wei, Cleaveland, Pappas, and
  Burdick]{dixit2023adaptive}
Anushri Dixit, Lars Lindemann, Skylar~X Wei, Matthew Cleaveland, George~J
  Pappas, and Joel~W Burdick.
\newblock Adaptive conformal prediction for motion planning among dynamic
  agents.
\newblock In \emph{Learning for Dynamics and Control Conference}, pages
  300--314. PMLR, 2023.

\bibitem[Ren et~al.(2023)Ren, Dixit, Bodrova, Singh, Tu, Brown, Xu, Takayama,
  Xia, Varley, et~al.]{ren2023robots}
Allen~Z Ren, Anushri Dixit, Alexandra Bodrova, Sumeet Singh, Stephen Tu, Noah
  Brown, Peng Xu, Leila Takayama, Fei Xia, Jake Varley, et~al.
\newblock Robots that ask for help: Uncertainty alignment for large language
  model planners.
\newblock \emph{arXiv preprint arXiv:2307.01928}, 2023.

\bibitem[Dixit et~al.(2024)Dixit, Mei, Booker, Storey-Matsutani, Ren, and
  Majumdar]{dixit2024perceive}
Anushri Dixit, Zhiting Mei, Meghan Booker, Mariko Storey-Matsutani, Allen~Z
  Ren, and Anirudha Majumdar.
\newblock Perceive with confidence: Statistical safety assurances for
  navigation with learning-based perception.
\newblock In \emph{Proceedings of the Conference on Robot Learning (CoRL)},
  2024.

\bibitem[Lindemann et~al.(2024)Lindemann, Zhao, Yu, Pappas, and
  Deshmukh]{lindemann2024formal}
Lars Lindemann, Yiqi Zhao, Xinyi Yu, George~J Pappas, and Jyotirmoy~V Deshmukh.
\newblock Formal verification and control with conformal prediction.
\newblock \emph{arXiv preprint arXiv:2409.00536}, 2024.

\bibitem[Betker et~al.(2023)Betker, Goh, Jing, Brooks, Wang, Li, Ouyang,
  Zhuang, Lee, Guo, et~al.]{betker2023improving}
James Betker, Gabriel Goh, Li~Jing, Tim Brooks, Jianfeng Wang, Linjie Li, Long
  Ouyang, Juntang Zhuang, Joyce Lee, Yufei Guo, et~al.
\newblock Improving image generation with better captions.
\newblock \emph{OpenAI}, 2023.

\bibitem[Nichol et~al.(2021)Nichol, Dhariwal, Ramesh, Shyam, Mishkin, McGrew,
  Sutskever, and Chen]{nichol2021glide}
Alex Nichol, Prafulla Dhariwal, Aditya Ramesh, Pranav Shyam, Pamela Mishkin,
  Bob McGrew, Ilya Sutskever, and Mark Chen.
\newblock Glide: Towards photorealistic image generation and editing with
  text-guided diffusion models.
\newblock \emph{arXiv preprint arXiv:2112.10741}, 2021.

\bibitem[Yu et~al.(2023{\natexlab{a}})Yu, Feng, Feng, Liu, Jin, Zeng, and
  Chen]{yu2023inpaint}
Tao Yu, Runseng Feng, Ruoyu Feng, Jinming Liu, Xin Jin, Wenjun Zeng, and Zhibo
  Chen.
\newblock Inpaint anything: Segment anything meets image inpainting.
\newblock \emph{arXiv preprint arXiv:2304.06790}, 2023{\natexlab{a}}.

\bibitem[Ling et~al.(2021)Ling, Kreis, Li, Kim, Torralba, and
  Fidler]{ling2021editgan}
Huan Ling, Karsten Kreis, Daiqing Li, Seung~Wook Kim, Antonio Torralba, and
  Sanja Fidler.
\newblock Editgan: High-precision semantic image editing.
\newblock \emph{Advances in Neural Information Processing Systems},
  34:\penalty0 16331--16345, 2021.

\bibitem[Zhu et~al.(2020)Zhu, Shen, Zhao, and Zhou]{zhu2020domain}
Jiapeng Zhu, Yujun Shen, Deli Zhao, and Bolei Zhou.
\newblock In-domain {GAN} inversion for real image editing.
\newblock In \emph{European conference on computer vision}, pages 592--608.
  Springer, 2020.

\bibitem[Chen et~al.(2023)Chen, Kiami, Gupta, and Kumar]{chen2023genaug}
Zoey Chen, Sho Kiami, Abhishek Gupta, and Vikash Kumar.
\newblock Genaug: Retargeting behaviors to unseen situations via generative
  augmentation.
\newblock \emph{arXiv preprint arXiv:2302.06671}, 2023.

\bibitem[Yu et~al.(2023{\natexlab{b}})Yu, Xiao, Stone, Tompson, Brohan, Wang,
  Singh, Tan, Peralta, Ichter, et~al.]{yu2023scaling}
Tianhe Yu, Ted Xiao, Austin Stone, Jonathan Tompson, Anthony Brohan, Su~Wang,
  Jaspiar Singh, Clayton Tan, Jodilyn Peralta, Brian Ichter, et~al.
\newblock Scaling robot learning with semantically imagined experience.
\newblock \emph{arXiv preprint arXiv:2302.11550}, 2023{\natexlab{b}}.

\bibitem[Bharadhwaj et~al.(2024)Bharadhwaj, Vakil, Sharma, Gupta, Tulsiani, and
  Kumar]{bharadhwaj2024roboagent}
Homanga Bharadhwaj, Jay Vakil, Mohit Sharma, Abhinav Gupta, Shubham Tulsiani,
  and Vikash Kumar.
\newblock Roboagent: Generalization and efficiency in robot manipulation via
  semantic augmentations and action chunking.
\newblock In \emph{Proceedings of the IEEE International Conference on Robotics
  and Automation (ICRA)}, pages 4788--4795, 2024.

\bibitem[Chen et~al.(2024{\natexlab{a}})Chen, Xu, Dharmarajan, Irshad, Cheng,
  Keutzer, Tomizuka, Vuong, and Goldberg]{chen2024rovi}
Lawrence~Yunliang Chen, Chenfeng Xu, Karthik Dharmarajan, Muhammad~Zubair
  Irshad, Richard Cheng, Kurt Keutzer, Masayoshi Tomizuka, Quan Vuong, and Ken
  Goldberg.
\newblock Ro{V}i-{A}ug: Robot and viewpoint augmentation for cross-embodiment
  robot learning.
\newblock \emph{arXiv preprint arXiv:2409.03403}, 2024{\natexlab{a}}.

\bibitem[Chen et~al.(2024{\natexlab{b}})Chen, Mandi, Bharadhwaj, Sharma, Song,
  Gupta, and Kumar]{chen2024semantically}
Zoey Chen, Zhao Mandi, Homanga Bharadhwaj, Mohit Sharma, Shuran Song, Abhishek
  Gupta, and Vikash Kumar.
\newblock Semantically controllable augmentations for generalizable robot
  learning.
\newblock \emph{The International Journal of Robotics Research}, page
  02783649241273686, 2024{\natexlab{b}}.

\bibitem[Black et~al.(2023)Black, Nakamoto, Atreya, Walke, Finn, Kumar, and
  Levine]{black2023zero}
Kevin Black, Mitsuhiko Nakamoto, Pranav Atreya, Homer Walke, Chelsea Finn,
  Aviral Kumar, and Sergey Levine.
\newblock Zero-shot robotic manipulation with pretrained image-editing
  diffusion models.
\newblock \emph{arXiv preprint arXiv:2310.10639}, 2023.

\bibitem[Shah et~al.(2023)Shah, Sridhar, Dashora, Stachowicz, Black, Hirose,
  and Levine]{shah2023vint}
Dhruv Shah, Ajay Sridhar, Nitish Dashora, Kyle Stachowicz, Kevin Black, Noriaki
  Hirose, and Sergey Levine.
\newblock Vint: A foundation model for visual navigation.
\newblock \emph{arXiv preprint arXiv:2306.14846}, 2023.

\bibitem[Hancock et~al.(2024)Hancock, Ren, and Majumdar]{hancock2024run}
Asher~J Hancock, Allen~Z Ren, and Anirudha Majumdar.
\newblock Run-time observation interventions make vision-language-action models
  more visually robust.
\newblock \emph{arXiv preprint arXiv:2410.01971}, 2024.

\bibitem[Precup(2000)]{precup2000eligibility}
Doina Precup.
\newblock Eligibility traces for off-policy policy evaluation.
\newblock \emph{Computer Science Department Faculty Publication Series},
  page~80, 2000.

\bibitem[Hallak and Mannor(2017)]{hallak2017consistent}
Assaf Hallak and Shie Mannor.
\newblock Consistent on-line off-policy evaluation.
\newblock In \emph{Proceedings of the International Conference on Machine
  Learning (ICML)}, pages 1372--1383, 2017.

\bibitem[Hanna et~al.(2017)Hanna, Stone, and Niekum]{hanna2017bootstrapping}
Josiah Hanna, Peter Stone, and Scott Niekum.
\newblock Bootstrapping with models: Confidence intervals for off-policy
  evaluation.
\newblock In \emph{Proceedings of the AAAI Conference on Artificial
  Intelligence}, volume~31, 2017.

\bibitem[Farajtabar et~al.(2018)Farajtabar, Chow, and
  Ghavamzadeh]{farajtabar2018more}
Mehrdad Farajtabar, Yinlam Chow, and Mohammad Ghavamzadeh.
\newblock More robust doubly robust off-policy evaluation.
\newblock In \emph{Proceedings of the International Conference on Machine
  Learning (ICML)}, pages 1447--1456, 2018.

\bibitem[Levine et~al.(2020)Levine, Kumar, Tucker, and Fu]{levine2020offline}
Sergey Levine, Aviral Kumar, George Tucker, and Justin Fu.
\newblock Offline reinforcement learning: Tutorial, review, and perspectives on
  open problems.
\newblock \emph{arXiv preprint arXiv:2005.01643}, 2020.

\bibitem[Luo et~al.(2024{\natexlab{b}})Luo, Sinha, Sun, Hindy, Zhao, Savarese,
  Schmerling, and Pavone]{luo2024online}
Rachel Luo, Rohan Sinha, Yixiao Sun, Ali Hindy, Shengjia Zhao, Silvio Savarese,
  Edward Schmerling, and Marco Pavone.
\newblock Online distribution shift detection via recency prediction.
\newblock In \emph{2024 IEEE International Conference on Robotics and
  Automation (ICRA)}, pages 16251--16263. IEEE, 2024{\natexlab{b}}.

\bibitem[Angelopoulos and Bates(2021)]{angelopoulos2021gentle}
Anastasios~N Angelopoulos and Stephen Bates.
\newblock A gentle introduction to conformal prediction and distribution-free
  uncertainty quantification.
\newblock \emph{arXiv preprint arXiv:2107.07511}, 2021.

\bibitem[Belkhale et~al.(2023)Belkhale, Cui, and Sadigh]{belkhale2023hydra}
Suneel Belkhale, Yuchen Cui, and Dorsa Sadigh.
\newblock Hydra: Hybrid robot actions for imitation learning.
\newblock In \emph{Proceedings of the Conference on Robot Learning}, pages
  2113--2133, 2023.

\bibitem[Zar(2005)]{zar2005spearman}
Jerrold~H Zar.
\newblock Spearman rank correlation.
\newblock \emph{Encyclopedia of Biostatistics}, 7, 2005.

\bibitem[Chen and Wang(2024)]{chen2024proedit}
Jun-Kun Chen and Yu-Xiong Wang.
\newblock Proedit: Simple progression is all you need for high-quality {3D}
  scene editing.
\newblock \emph{arXiv preprint arXiv:2411.05006}, 2024.

\bibitem[Bao et~al.(2024)Bao, Ding, Huo, Liu, Li, Li, Gao, and Luo]{bao20243d}
Yanqi Bao, Tianyu Ding, Jing Huo, Yaoli Liu, Yuxin Li, Wenbin Li, Yang Gao, and
  Jiebo Luo.
\newblock 3d {G}aussian splatting: Survey, technologies, challenges, and
  opportunities.
\newblock \emph{arXiv preprint arXiv:2407.17418}, 2024.

\bibitem[Xu et~al.()Xu, Nguyen, Miller, Lee, Shah, Ambrus, Nishimura, and
  Itkina]{xuuncertainty}
Chen Xu, Tony~Khuong Nguyen, Patrick Miller, Robert Lee, Paarth Shah,
  Rares~Andrei Ambrus, Haruki Nishimura, and Masha Itkina.
\newblock Uncertainty-aware failure detection for imitation learning robot
  policies.
\newblock In \emph{CoRL Workshop on Safe and Robust Robot Learning for
  Operation in the Real World}.

\bibitem[Dosovitskiy et~al.(2020)Dosovitskiy, Beyer, Kolesnikov, Weissenborn,
  Zhai, Unterthiner, Dehghani, Minderer, Heigold, Gelly, Uszkoreit, and
  Houlsby]{vit}
Alexey Dosovitskiy, Lucas Beyer, Alexander Kolesnikov, Dirk Weissenborn,
  Xiaohua Zhai, Thomas Unterthiner, Mostafa Dehghani, Matthias Minderer, Georg
  Heigold, Sylvain Gelly, Jakob Uszkoreit, and Neil Houlsby.
\newblock An image is worth 16x16 words: Transformers for image recognition at
  scale.
\newblock \emph{arXiv preprint arXiv:2010.11929}, 2020.

\bibitem[Ryoo et~al.(2021)Ryoo, Piergiovanni, Arnab, Dehghani, and
  Angelova]{ryoo2021tokenlearner}
Michael Ryoo, AJ~Piergiovanni, Anurag Arnab, Mostafa Dehghani, and Anelia
  Angelova.
\newblock Tokenlearner: Adaptive space-time tokenization for videos.
\newblock \emph{Advances in neural information processing systems},
  34:\penalty0 12786--12797, 2021.

\bibitem[Peebles and Xie(2023)]{diffusion-transformer}
William Peebles and Saining Xie.
\newblock Scalable diffusion models with transformers.
\newblock In \emph{Proceedings of the IEEE/CVF International Conference on
  Computer Vision}, pages 4195--4205, 2023.

\bibitem[Reuss et~al.(2023)Reuss, Li, Jia, and Lioutikov]{reuss2023goal}
Moritz Reuss, Maximilian Li, Xiaogang Jia, and Rudolf Lioutikov.
\newblock Goal-conditioned imitation learning using score-based diffusion
  policies.
\newblock \emph{arXiv preprint arXiv:2304.02532}, 2023.

\bibitem[Karras et~al.(2022)Karras, Aittala, Aila, and
  Laine]{karras2022elucidating}
Tero Karras, Miika Aittala, Timo Aila, and Samuli Laine.
\newblock Elucidating the design space of diffusion-based generative models.
\newblock \emph{Advances in neural information processing systems},
  35:\penalty0 26565--26577, 2022.

\end{thebibliography}

\newpage
\appendix
\section{Image Editing: Examples and Prompts}
\label{app:image editing}

\begin{figure}[htbp]
    \centering
    \begin{subfigure}[b]{0.49\textwidth} 
        \includegraphics[width=\textwidth]{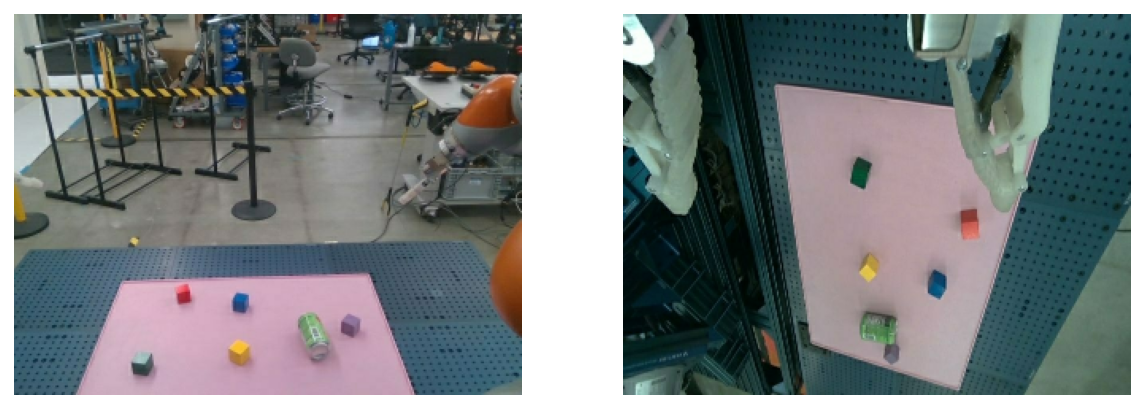} 
        \caption{Nominal overhead (left) and wrist (right) cameras.}
        \label{fig:sub1}
    \end{subfigure}
    \hfill
    \begin{subfigure}[b]{0.49\textwidth} 
        \includegraphics[width=\textwidth]{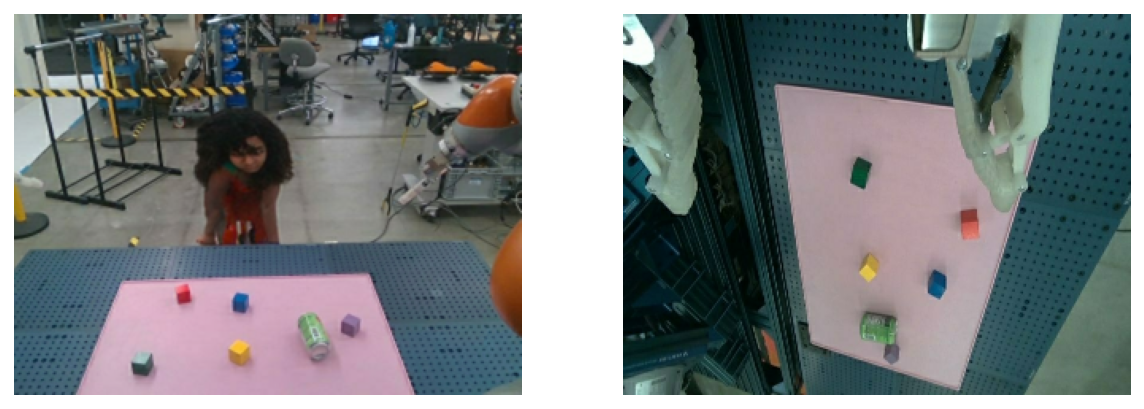} 
        \caption{Distractor: person.}
        \label{fig:sub2}
    \end{subfigure}

    \bigskip 

    \begin{subfigure}[b]{0.49\textwidth} 
        \includegraphics[width=\textwidth]{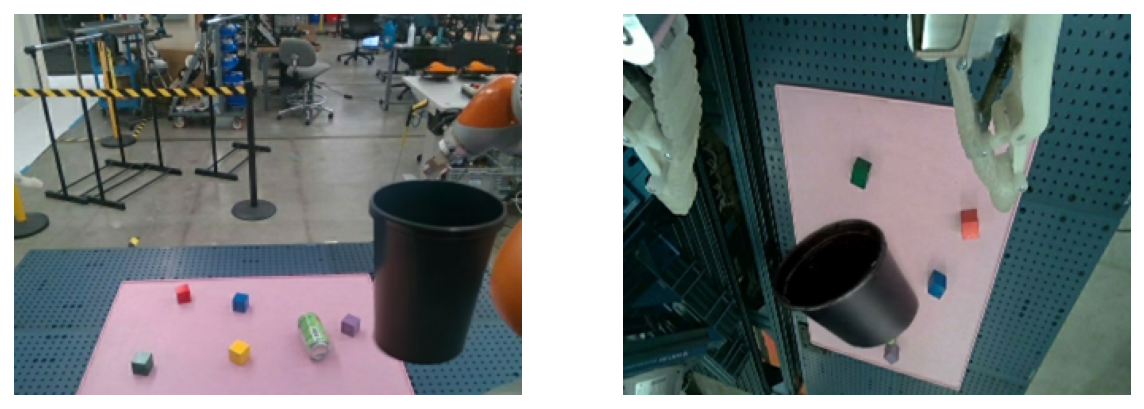} 
        \caption{Distractor: trash can.}
        \label{fig:sub3}
    \end{subfigure}
    \hfill
    \begin{subfigure}[b]{0.49\textwidth} 
        \includegraphics[width=\textwidth]{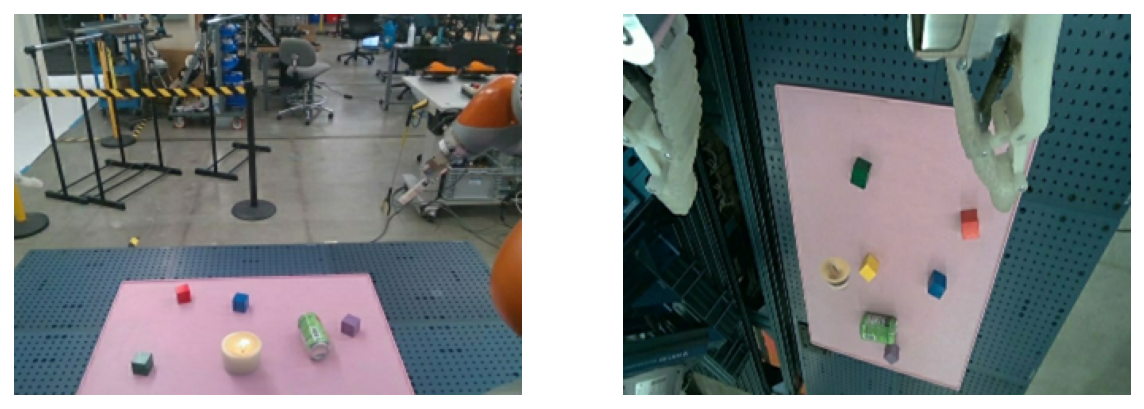} 
        \caption{Distractor: candle.}
        \label{fig:sub4}
    \end{subfigure}

    \bigskip 

    \begin{subfigure}[b]{0.49\textwidth} 
        \includegraphics[width=\textwidth]{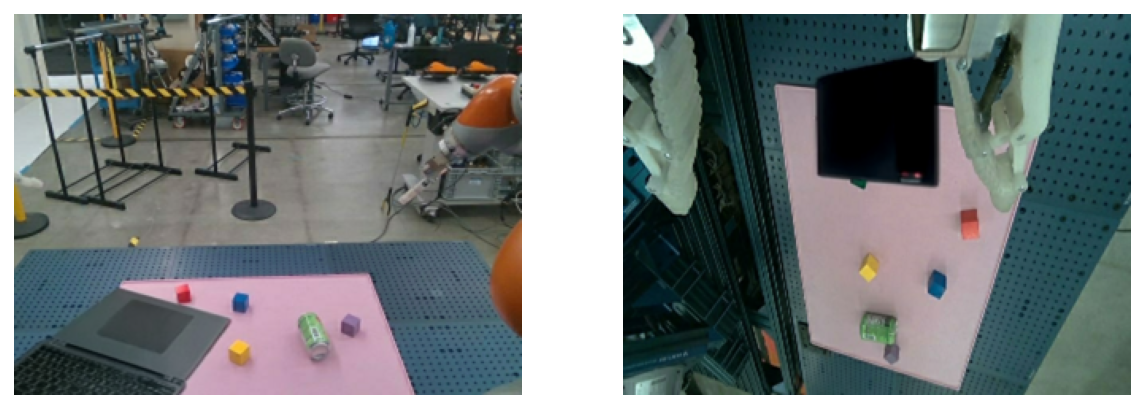} 
        \caption{Distractor: laptop.}
        \label{fig:sub5}
    \end{subfigure}
    \hfill
    \begin{subfigure}[b]{0.49\textwidth} 
        \includegraphics[width=\textwidth]{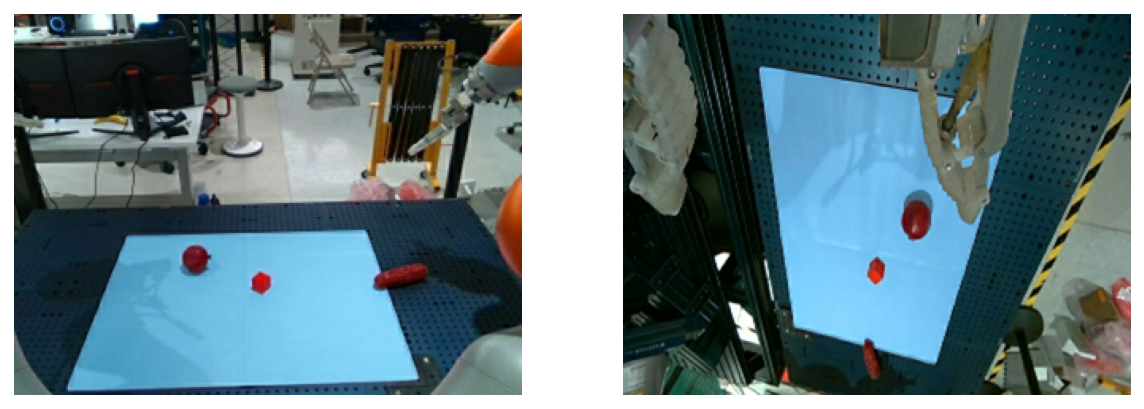} 
        \caption{Background: blue.}
        \label{fig:sub6}
    \end{subfigure}

    \bigskip 

    \begin{subfigure}[b]{0.49\textwidth} 
        \includegraphics[width=\textwidth]{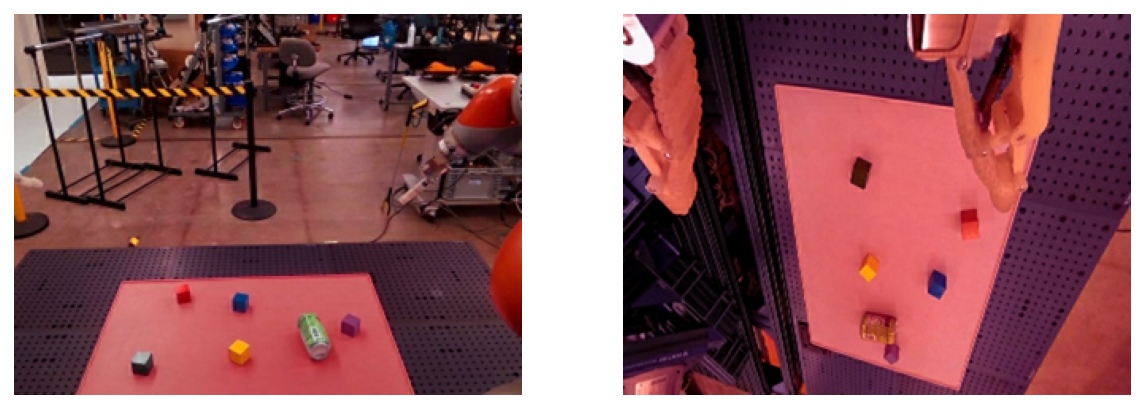} 
        \caption{Lighting: red.}
        \label{fig:sub7}
    \end{subfigure}
    \hfill
    \begin{subfigure}[b]{0.49\textwidth} 
        \includegraphics[width=\textwidth]{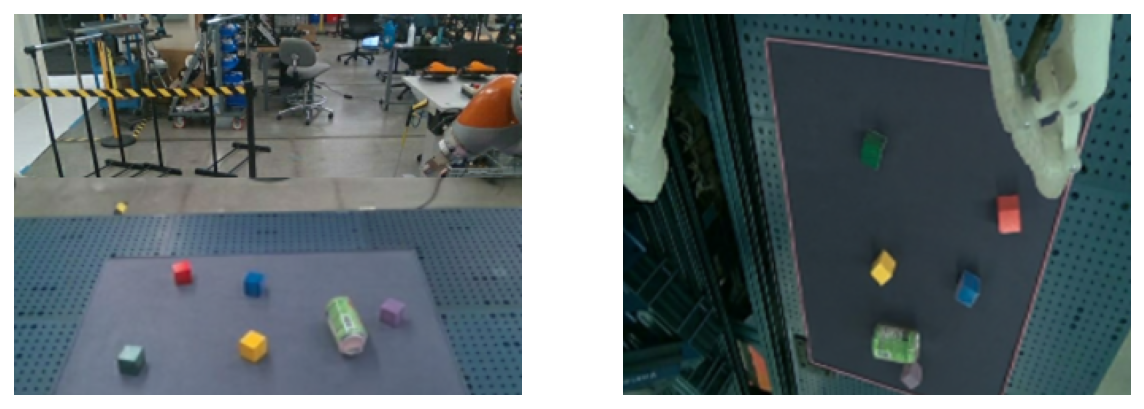} 
        \caption{Table height (changed color followed by zoom).}
        \label{fig:sub8}
    \end{subfigure}

    \caption{Examples of different edits applied to nominal overhead and wrist camera observations.}
    \label{fig:example edits}
\end{figure}

Examples of different edits applied to both the overhead camera and the wrist camera are shown in Figure~\ref{fig:example edits}. Below, we provide complete prompts used to generate the edited observations for each environmental condition. 

\begin{tcolorbox}[
    colback=myboxcolor,
    colframe=myboxcolor,
    breakable,
    boxrule=0mm,
    arc=0mm,
    boxsep=0mm,
    left=3mm,
    right=3mm,
    top=3mm,
    bottom=3mm,
    ]
{\Large {\bf Full prompts for edits}} 

\vspace{10pt}

{\bf Person:} \\
Add a person to the image. Specifically, add a person behind the blue platform, realistically interacting with the platform and fitting seamlessly into the existing environment. Preserve all other aspects of the image, including the different objects on the mat, other background  elements, and the overall composition.  The lighting should remain consistent.  The new person should be realistically rendered with all details of the person including their face, clothing, and any other visible parts shown in exquisite clarity and detail. Only the person should be added. \\
{\bf Large distractor (e.g., trash cans):} \\
Add a large $<$\texttt{target color}$>$ $<$\texttt{target object}$>$ at the edge of the pink mat, so that it doesn't modify or occlude any of the objects on the pink mat. Specifically, add a $<$\texttt{target color}$>$ $<$\texttt{target object}$>$ that is larger than any objects at the edge of the pink mat, fitting realistically and seamlessly into the existing scene. Preserve all details of the objects on the mat, their poses, and the overall composition of the image. The $<$\texttt{target color}$>$ $<$\texttt{target object}$>$ should be realistically and exquisitely rendered and should not occlude any of the objects on the pink mat. The lighting should remain consistent. Only the $<$\texttt{target color}$>$ $<$\texttt{target object}$>$ at the edge of the pink mat should be added. \\
{\bf Small distractor (e.g., candle):} \\
Modify image $<$\texttt{image}$>$ as described below: Add a small scented candle on the pink mat, so that it doesn't modify or occlude any of the objects on the mat. Specifically, add a scented candle with roughly the same size as the objects on the pink mat, fitting realistically and seamlessly into the existing scene. Preserve all details of the composition of the image. The scented candle should be realistically and exquisitely rendered and should not occlude any of the objects on the pink mat. The lighting should remain consistent. Only the scented candle should be added. \\
 {\bf Background color:} \\
Modify image $<$\texttt{image}$>$ as described below: change the color of the pink mat that objects are on to $<$\texttt{target color}$>$. Preserve the different objects on the mat, and all other aspects of the image including the lighting and the overall composition. \\
{\bf Lighting (overhead camera):} \\
Modify image $<$\texttt{image}$>$ as described below: Colorize the bottom half of the image with an extremely intense $<$\texttt{target color}$>$ hue. Preserve the existing composition, details, and textures of the objects in the scene, including the ones on the pink mat and the background.  Only the shadows and color palette should be altered to reflect an extremely intense $<$\texttt{target color}$>$ light, maintaining the style of the original image. The overall lighting should remain consistent, with shadows and highlights adjusted to match the new color palette. Make sure that the hue for the bottom half of the image is changed to intense $<$\texttt{target color}$>$, including for the objects on the table. \\
{\bf Lighting (wrist camera):} \\
Modify image $<$\texttt{image}$>$ as described below: Colorize the entire image with an extremely intense $<$\texttt{target color}$>$ color tone. Preserve the existing composition, details, and textures of the objects in the scene, including the ones on the pink mat and the background.  Only the shadows and color palette should be altered to reflect an extremely intense $<$\texttt{target color}$>$ light, maintaining the style of the original image. The overall lighting should remain consistent, with shadows and highlights adjusted to match the new color palette. Make sure that the color for the entire image is changed to intense $<$\texttt{target color}$>$. \\
{\bf Table height:} \\
Change the color of the pink mat to $<$\texttt{target color}$>$. Preserve all other aspects of the image, including the different objects on the mat, the lighting, and the overall composition. Only the color of the pink mat should be altered to $<$\texttt{target color}$>$, maintaining its shape, size, and position. [We then apply a zoom to the portion of the image that contains the table in order to simulate a change in the height of the table.
\end{tcolorbox}

\section{Filtering Edits with a Vision-Language Model}
\label{app:vlm filter}

For each nominal observation, we generate a batch of four candidate edited observations via the image editing model. We then use a vision-language model (VLM) in order to judge if any of the options accurately reflect the desired change; if so, the VLM is tasked with choosing the best one (if not, we simply discard the observation from our set). The full prompt for the VLM --- which involves chain-of-thought reasoning --- is provided below. We use the Gemini Pro 1.5 VLM~\cite{team2023gemini} for our experiments.

\begin{tcolorbox}[
    colback=myboxcolor,
    colframe=myboxcolor,
    breakable,
    boxrule=0mm,
    arc=0mm,
    boxsep=0mm,
    left=3mm,
    right=3mm,
    top=3mm,
    bottom=3mm,
    ]
{\Large {\bf Prompt for filtering edits with a VLM}} 

\vspace{10pt}

Here is the original image I have: $<$\texttt{original image}$>$. Do any of Image~0: $<$\texttt{Image 0}$>$, Image~1: $<$\texttt{Image 1}$>$, Image~2: $<$\texttt{Image 2}$>$, or Image~3 $<$\texttt{Image 3}$>$ accurately reflect an edited version of the original image with the instruction ``$<$\texttt{short edit instruction}$>$"? Give your reasoning and then answer with a True or False. If True, provide the index (0,1,2,3) of the best image.
\end{tcolorbox}

The variable $<$\texttt{short edit instruction}$>$ contains a shortened version (e.g., ``Change the color of the pink mat to $<$\texttt{target color}$>$") of the full prompt provided to the image editing model. We find that providing the full prompt (instead of a shortened version) can lead the VLM to be overly critical and filter out many acceptable edits.

\section{Training and Policy Details}
\label{app:policy}

\subsection{Training Data Collection}

{\bf Training data collection.} For training our policies, we collect 3K+ demonstrations for grasping tasks on the hardware. Specifically, we use trajectory optimization-based motion planners to automatically collect a large set of training data. Our data collection pipeline uses the overhead camera to obtain a 3D point cloud of the scene. We segment the point cloud into multiple objects and randomly choose different objects to pick using the left arm. We use automated success detection to segment these trajectories. For each episode, we further automatically annotate keypoints for the object the policy should grasp; these are used as additional context for the robot policy in addition to camera and proprioceptive observations. All demonstrations are collected in nominal conditions, i.e., with fixed lighting, with a fixed pink background on a table, and an object set that consists of blocks, plush toys, small cans, and artificial fruits.

\subsection{Hybrid Policy Architecture}
\begin{figure*}[t]
    \centering
    \includegraphics[width=0.95\textwidth]{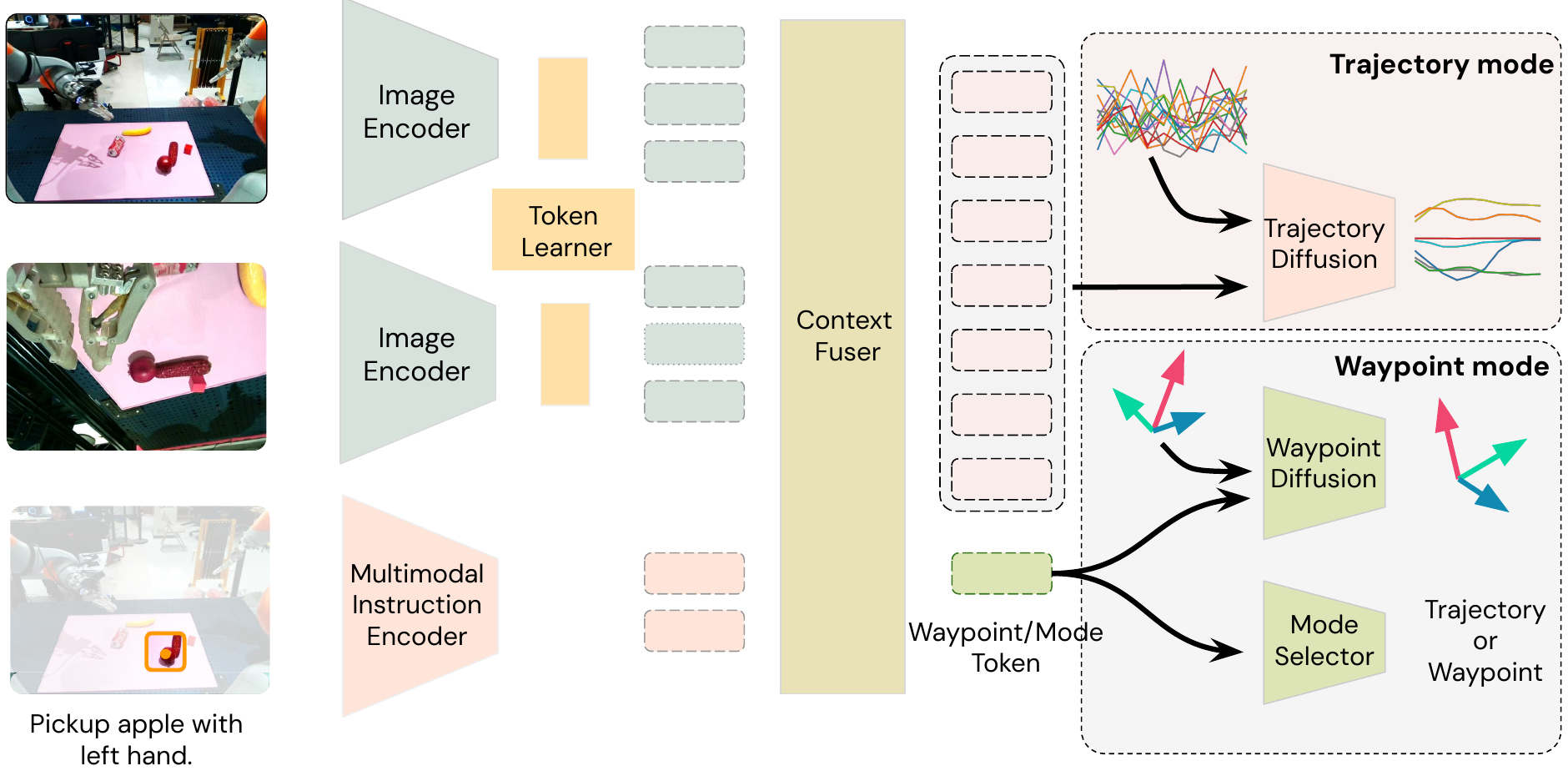}
    \caption{The policy architecture used for two different --- hybrid and diffusion --- policy implementation. Our unified architecture consists of a trajectory mode which predicts the continuous joint space actions and a waypoint mode which predicts a single SE(3) waypoint that the arm should reach to. We use two different policies. 1. \emph{hybrid policy} uses both the trajectory and waypoint mode and selects between them to execute the action. 2. \emph{\expandafter\nonhybridpolicy} only uses the trajectory mode and directly predicts the joint space trajectory for the robot to follow.}
    \label{fig:architecture}
\end{figure*}

{\bf Hybrid policy.} We consider two policies that vary significantly in their overall architecture. The first policy $\pi_\text{hyb}$ uses a hybrid policy architecture inspired by \cite{belkhale2023hydra}, which aims to utilize the benefits of trajectory optimization-based approaches for free space planning together with the reactive nature of closed-loop visuomotor diffusion policies. We achieve this by using two separate heads in our policy architecture (see Fig.~\ref{fig:architecture}), where each head represents an \emph{action mode}. These two different action modes correspond to:
\begin{enumerate}
    \item a waypoint action mode which outputs a single waypoint $(w \in SE(3))$, and
    \item a trajectory action mode which outputs a dense sequence of robot joint angles $(q_i \in \mathbb{R}^{14})$.
\end{enumerate}
In addition to these policy heads we also output a mode selection scalar which defines which action mode should be executed at any given time. In order to execute the waypoint action we use a trajectory optimization approach based on sequential quadratic programming (SQP), and execute the output trajectory for a fixed number of steps before re-querying the policy. By contrast, in order to execute the trajectory action we simply interpolate through the joint commands outputted by the network. Importantly, during training both policy heads are trained \emph{simultaneously}, i.e., each input data item is labelled with a waypoint action (extracted using an object closeness heuristic) and a dense trajectory action (which we directly extract from the robot logs). We supervise the mode selection scalar to output the waypoint action mode when the arm is far away from any object and the trajectory action mode in all other scenarios. 

{\bf Vision encoder.} Our policy architecture uses pre-trained ViT \cite{vit} encoders to encode the image observations from each image. We use separate models for each camera observation (overhead and wrist). We reduce the number of tokens from each ViT using a TokenLearner layer \cite{ryoo2021tokenlearner}. We encode proprioceptive features using a multi-layer perceptron (MLP) with a single hidden layer.

{\bf Instruction encoder.} The robot is instructed to grasp a target object using semantic keypoints. Specifically, we extract a small patch $(64 \times 64)$ from the overhead camera view around a keypoint that is selected by the robot operator. We encode this patch using a small coordinate convolution-based neural network.
Since we train a multi-skill policy we encode the skill that the robot needs to perform using a continuous embedding.
The semantic keypoint representation is concatenated with the skill embedding to form the instruction tokens.

{\bf Context Fuser.} The observation tokens, the instruction tokens and proprioceptive tokens are fused together using a context fuser 
which uses a stack of self-attention based transformer layers. We also additionally add a readout token, which we refer to as the waypoint-mode token.
At the end of the context fuser we get a set of fused observation-instruction embeddings as well as the embedding for the readout token.  The observation-instruction embeddings are used to predict the trajectory and thus passed into the trajectory diffusion transformer.
Alternatively, the waypoint-mode embedding is used by the waypoint diffusion transformer to predict the $SE(3)$ waypoint as well as to
predict the current mode for the robot. The observation-instruction embeddings are used by \redit~for anomaly detection. 

{\bf Diffusion.} For both trajectory diffusion and waypoint diffusion we use a Transformer decoder-based denoiser \cite{diffusion-transformer}.
The denoiser takes as input noisy action embeddings together with a diffusion timestep embedding.
These noisy actions and timestep embeddings cross-attend to the context embeddings (either the context tokens for trajectory diffusion or waypoint embedding for waypoint prediction).
After multiple layers of alternating between self-attention and cross-attention the diffusion transformer outputs the denoised trajectory or waypoint action (as desired).

\subsection{Diffusion Policy.}
Our \nonhybridpolicy architecture $\pi_\text{dfn}$ uses a standard diffusion policy \cite{chi2023diffusion, reuss2023goal}
to directly output the joint angles to control the robot.
Our base architecture is similar to $\pi_\text{hyb}$ (described above) wherein we only use the trajectory mode, i.e.,
only the trajectory diffusion head is used to predict robot trajectories.
The rest of the architecture including the vision encoders and the multi-modal instruction encoder are common between
$\pi_\text{dfn}$ and $\pi_\text{hyb}$.
However, unlike $\pi_\text{hyb}$, $\pi_\text{dfn}$ does not include a readout token (waypoint/mode token) within the context fuser.

\begin{table}[]
\centering
\resizebox{0.5\textwidth}{!}{%
\begin{tabular}{@{}ll@{}}
\toprule
Hyperparam                  & Value                 \\ \midrule
train steps                 & 500K                  \\
optimizer                   & AdamW                 \\
warmup                      & linear upto 10K steps \\
learning rate               & 1e-4                  \\
learnign rate decay         & constant              \\
weight decay                & 1e-4                  \\
trajectory (action) horizon & 10                    \\ \bottomrule
\end{tabular}%
}
\caption{Hyperparameters used to train the different policies used.}
\label{tab:train-hparams-table}
\end{table}

\subsection{Training and Inference Details}
\label{app:train-details}

Table~\ref{tab:train-hparams-table} shows the common set of hyper-parameters used to train each of our policies.
We use a batch size of 256 during training.
As shown in Figure~\ref{fig:architecture} for the high dimensional image observations we use an additional token learner to reduce the number of image tokens. We use 128 tokens for each image observation.
For our diffusion model we use a continuous time implementation based on \cite{karras2022elucidating}. Similar to \cite{karras2022elucidating} we use a second order Heun solver to solve the flow ODE. 
We use 30 ODE steps during inference. As shown in Table~\ref{tab:train-hparams-table}, we use an action horizon of 10.
Since we collect our training data at 10Hz this corresponds to 1 second of robot motion. During inference, we open-loop rollout
entire 10 steps before querying the policy again.
During evaluation we use a maximum of 30 policy steps before we stop policy evaluation.
For our targeted data collection experiment Section~\ref{subsec:targeted-data-collection}, we use a much smaller learning rate of 
$5e-6$ and a linear warmup of 4K steps. We finetune the policy for a total 20K steps.

\section{Ablations: Score Function and Size of Nominal Rerence Set}
\label{app:ablations}

\subsection{Ablations for \redit}

\redit~uses an anomaly score function $s_\pi(o, S_\text{nom})$ that computes the mean of the $k$-nearest neighbor cosine distances in policy embedding space. The set $S_\text{nom}$ consists of embeddings computed for a random subset of the training data. Intuitively, this anomaly score quantifies how dissimilar a given observation is compared to similar training observations from the perspective of the policy. In the tables below, we provide results from varying the size of $S_\text{nom}$ and the value of $k$ for each policy. Each table compares the predictions made by \redit~with the actual (empirically measured) performance by computing the Spearman rank correlation and the average prediction error, as described in Sec.~\ref{sec:redit experiments}. Generally, we find that predictions for $\pi_\text{hyb}$ remain accurate when varying $|S_\text{nom}|$ with small $k$, while predictions for $\pi_\text{dfn}$ (which has a significantly higher dimensional embedding space) benefit from either having a smaller value of $|S_\text{nom}|$ or larger values of $k$. 

\vspace{10pt} 

{\bf Hybrid policy} $\pi_\text{hyb}$

\vspace{10pt} 

\begin{tabular}{c || *{4}{c}}
  \hline
  $|S_\text{nom}| = 3000$ & $k=1$ & $k=5$ & $k=10$   \\
  \hline
  \hline
  Spearman $\rho$ ({\bf $\uparrow$}) & 0.72 & {\bf 0.78} & 0.78 \\
  \hline
  Av. pred. err. ({\bf $\downarrow$}) & 0.12 & {\bf 0.1} & 0.12  \\
  \hline
\end{tabular}

\vspace{10pt} 

\begin{tabular}{c || *{4}{c}}
  \hline
  $|S_\text{nom}| = 2000$ & $k=1$ & $k=5$ & $k=10$   \\
  \hline
  \hline
  Spearman $\rho$ ({\bf $\uparrow$}) & 0.72 & {\bf 0.79} & 0.68 \\
  \hline
  Av. pred. err. ({\bf $\downarrow$}) & 0.12 & {\bf 0.12} & 0.13  \\
  \hline
\end{tabular}

\vspace{10pt}

\begin{tabular}{c || *{3}{c}}
  \hline
  $|S_\text{nom}| = 1000$ & $k=1$ & $k=5$ & $k=10$  \\
  \hline
  \hline
  Spearman $\rho$ ({\bf $\uparrow$}) & {\bf 0.76} & 0.65 & 0.63 \\
  \hline
  Av. pred. err. ({\bf $\downarrow$}) & {\bf 0.12} & 0.13 & 0.17  \\
  \hline
\end{tabular}

\vspace{10pt}

\begin{tabular}{c || *{3}{c}}
  \hline
  $|S_\text{nom}| = 500$ & $k=1$ & $k=5$ & $k=10$  \\
  \hline
  \hline
  Spearman $\rho$ ({\bf $\uparrow$}) & {\bf 0.69} & 0.63 & 0.56 \\
  \hline
  Av. pred. err. ({\bf $\downarrow$}) & {\bf 0.12} & 0.16 & 0.19  \\
  \hline
\end{tabular}

\vspace{10pt}

\begin{tabular}{c || *{3}{c}}
  \hline
  $|S_\text{nom}| = 200$ & $k=1$ & $k=5$ & $k=10$  \\
  \hline
  \hline
  Spearman $\rho$ ({\bf $\uparrow$}) & {\bf 0.72} & 0.65 & 0.49 \\
  \hline
  Av. pred. err. ({\bf $\downarrow$}) & {\bf 0.14} & 0.18 & 0.23  \\
  \hline
\end{tabular}

\vspace{10pt}

{\bf Vanilla diffusion policy} $\pi_\text{dfn}$

\vspace{10pt}

\begin{tabular}{c || *{7}{c}}
  \hline
  $|S_\text{nom}| = 3000$ & $k=1$ & $k=5$ & $k=10$ & $k=25$ & $k=50$ & $k=100$ & $k=200$  \\
  \hline
  \hline
  Spearman $\rho$ ({\bf $\uparrow$}) & 0.59 & 0.52 &  0.56 &  0.66 & 0.59 & {\bf 0.67} & 0.66 \\
  \hline
  Av. pred. err. ({\bf $\downarrow$}) & 0.22 & 0.21 & 0.21 &  0.20 & 0.20 & {\bf 0.19} & 0.20 \\
  \hline
\end{tabular}

\vspace{10pt}

\begin{tabular}{c || *{7}{c}}
  \hline
  $|S_\text{nom}| = 2000$ & $k=1$ & $k=5$ & $k=10$ & $k=25$ & $k=50$ & $k=100$ & $k=250$  \\
  \hline
  \hline
  Spearman $\rho$ ({\bf $\uparrow$}) & 0.55 & 0.52 &  0.53 &  0.64 & 0.64 & {\bf 0.69} & 0.17 \\
  \hline
  Av. pred. err. ({\bf $\downarrow$}) & 0.21 & 0.20 & 0.20 &  0.20 & 0.20 & 0.20 & 0.25 \\
  \hline
\end{tabular}

\vspace{10pt}

\begin{tabular}{c || *{4}{c}}
  \hline
  $|S_\text{nom}| = 1000$ & $k=1$ & $k=5$ & $k=10$ & $k=25$  \\
  \hline
  \hline
  Spearman $\rho$ ({\bf $\uparrow$}) & {\bf 0.63} & 0.52 &  0.60 &  0.59 \\
  \hline
  Av. pred. err. ({\bf $\downarrow$}) & 0.21 & 0.20 & {\bf 0.19} &  0.20 \\
  \hline
\end{tabular}

\vspace{10pt}

\begin{tabular}{c || *{3}{c}}
  \hline
  $|S_\text{nom}| = 500$ & $k=1$ & $k=5$ & $k=10$  \\
  \hline
  \hline
  Spearman $\rho$ ({\bf $\uparrow$}) & 0.58 & 0.66 & {\bf 0.71} \\
  \hline
  Av. pred. err. ({\bf $\downarrow$}) & 0.21 & 0.19 & {\bf 0.19}  \\
  \hline
\end{tabular}

\vspace{10pt}

\begin{tabular}{c || *{3}{c}}
  \hline
  $|S_\text{nom}| = 200$ & $k=1$ & $k=5$ & $k=10$  \\
  \hline
  \hline
  Spearman $\rho$ ({\bf $\uparrow$}) & 0.61 & 0.64 & {\bf 0.75} \\
  \hline
  Av. pred. err. ({\bf $\downarrow$}) & 0.19 & 0.20 & {\bf 0.19}  \\
  \hline
\end{tabular}

\vspace{10pt}

\section{Predicting Performance From Anomalies}
\label{app:anomaly-to-real}

\begin{figure*}[h]
    \centering
    \includegraphics[width=0.8\textwidth]{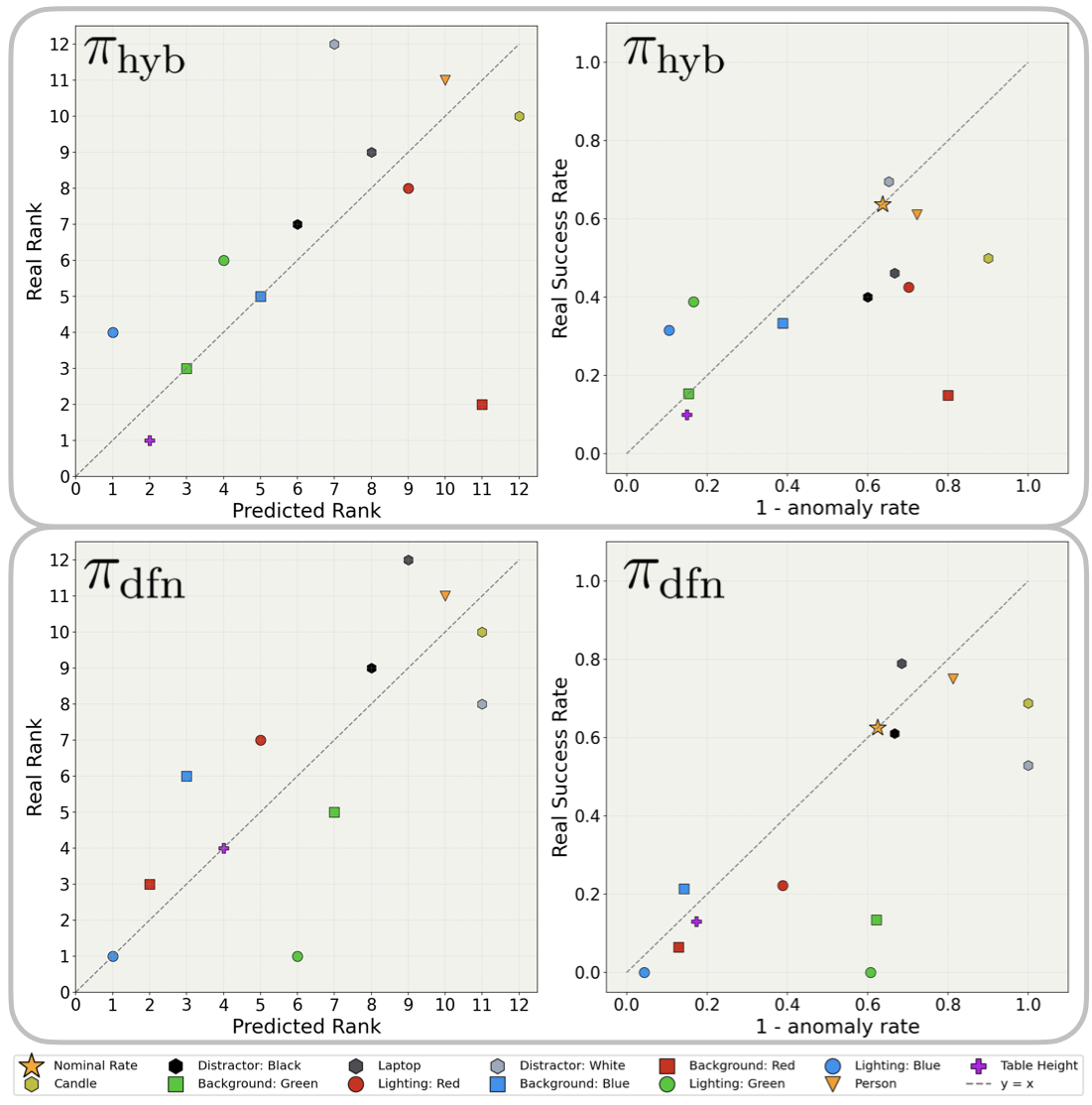}
    \caption{Evaluating predictions made from anomaly rates computed using real observations.}
    \label{fig:anomaly-vs-real}
    \vspace{-10pt}
\end{figure*}

Fig.~\ref{fig:anomaly-vs-real} compares the true (estimated) rankings of different environmental factors and success rates with rankings and success rates predicted by executing the anomaly detector on $\sim20$ real observations collected from each of the twelve off-nominal settings. Specifically, predicted success rates for each factor are computed as  $R_{f, \text{anom}}^\pi := 1 - \alpha_{f, \text{real}}^\pi$. For anomaly detection, we use $k=10, |S_\text{nom}| = 3000$ for $\pi_\text{hyb}$ and $k=10, |S_\text{nom}| = 200$ for $\pi_\text{dfn}$. The anomaly threshold for $\pi_\text{hyb}$ is computed using conformal prediction as described in Sec.~\ref{sec:anomaly detection} in order to bound the anomaly rate in nominal conditions to $1 - R_\text{nom}^{\pi_\text{hyb}}$. For $\pi_\text{dfn}$, we found this procedure to yield an anomaly threshold that is too conservative (i.e., flagging most observations in the different off-nominal scenarios as anomalous). This sensitivity may be due to the relatively small number $n_\text{val} = 70$ of nominal observations we used to compute the anomaly threshold and the very high dimensionality of the embedding space ($\mathbb{R}^{515\times513}$) of $\pi_\text{dfn}$. In order to correct for this, we computed the anomaly threshold with a slightly higher estimate of the nominal success rate (0.8 vs. 0.65), i.e., using conformal prediction to bound the anomaly rate in nominal conditions to $1 - 0.8$ rather than $1-0.65$. Fig.~\ref{fig:anomaly-vs-real} shows a strong correlation between predicted and realized performance. We note that the predictions $R_{f, \text{anom}}^\pi$ are made using $5\times$ fewer observations than predictions from the full \redit pipeline ($\sim 20$ real observations vs. $100$ edited observations), thus making them significantly more susceptible to noise. 

\end{document}